\documentclass{article}

 \usepackage[preprint]{neurips_2026}

\usepackage[utf8]{inputenc} 
\usepackage[T1]{fontenc}    
\usepackage{hyperref}       
\usepackage{url}            
\usepackage{booktabs}       
\usepackage{amsfonts}       
\usepackage{nicefrac}       
\usepackage{microtype}      

\hypersetup{hidelinks}
\usepackage{amsmath}
\usepackage{amssymb}
\usepackage{mathtools}
\usepackage{amsthm}
\usepackage{subcaption}
\usepackage{fvextra}
\usepackage{graphicx}
\usepackage{subcaption}
\usepackage{multirow}
\usepackage[table,xcdraw]{xcolor}
\usepackage[ruled,vlined]{algorithm2e}
\usepackage{enumitem}
\usepackage{placeins}
\usepackage{wrapfig}
\usepackage{siunitx}

\newcommand{\best}[1]{\bfseries #1}

\title{FlowCompile: An Optimizing Compiler for \\Structured LLM Workflows}

\author{%
  Junyan Li \\
  UMass Amherst \\
  \texttt{junyanli@umass.edu} \\
  \And
  Zhang-Wei Hong \\
  MIT-IBM Watson AI Lab \\
  \texttt{zwhong@mit.edu} \\
  \And
  Maohao Shen \\
  MIT \\
  \texttt{maohao@mit.edu} \\
  \And
  Yang Zhang \\
  MIT-IBM Watson AI Lab \\
  \texttt{yang.zhang2@ibm.com} \\
  \And
  Chuang Gan \\
  UMass Amherst \\
  MIT-IBM Watson AI Lab \\
  \texttt{chuangg@cs.umass.edu} \\
}

\begin{document}

\maketitle

\begin{abstract}

Structured LLM workflows, in which specialized LLM sub-agents are executed according to a predefined execution graph, have become a powerful abstraction for solving complex tasks. Optimizing such workflows, \emph{i.e.}, selecting configurations for each sub-agent to balance accuracy and latency, is fundamentally challenging due to the combinatorial design space over model choices, reasoning budgets, and workflow structures. Existing cost-aware methods largely treat workflow optimization as a routing problem, selecting a configuration at inference time for each query according to the accuracy--latency objective specified during training. We argue that, beyond runtime routing, structured LLM workflows can also be optimized from a \emph{compilation} perspective: before deployment, the system can globally explore the workflow design space and construct a reusable set of workflow-level configurations spanning diverse accuracy--latency trade-offs. Drawing inspiration from machine learning compilers, we introduce \textbf{FlowCompile}, a structured LLM workflow compiler that performs compile-time design space exploration to identify such a high-quality, reusable trade-off set. FlowCompile decomposes a workflow into sub-agents, profiles each sub-agent under diverse configurations, and composes these measurements through a structure-aware proxy to estimate workflow-level accuracy and latency. It then identifies a diverse set of high-quality configurations in a single compile-time pass, without retraining or online adaptation. Experiments across diverse workflows and challenging benchmarks show that FlowCompile consistently outperforms heuristically optimized workflow configurations and routing-based baselines by a large margin, delivering up to $\mathbf{6.4\times}$ speedup while maintaining strong task performance. Furthermore, the compiled configuration set serves as a reusable optimization artifact, enabling flexible deployment under varying runtime preferences and naturally supporting downstream selection or routing for additional gains. Code is released at: \url{https://github.com/UMass-Embodied-AGI/FlowCompile}.

\end{abstract}

\section{Introduction}
\label{sec:intro}

Recent advances in machine learning compilers, such as TVM~\citep{chen2018tvm}, Glow~\citep{rotem2018glow}, and XLA~\citep{xla}, have enabled efficient optimization of neural networks, including large language models (LLMs). TVM, in particular, illustrates a compiler-based approach that statically analyzes computation graphs, profiles low-level operator performance, and searches over execution configurations to optimize a target computation for a given deployment setting. It provides a scalable framework for exploring large design spaces and identifying efficient configurations.

As LLM systems evolve beyond single-model inference, they are increasingly instantiated as \textit{structured LLM workflows} composed of multiple specialized LLM sub-agents~\citep{zhang2024aflow,hu2024automated}. A structured LLM workflow connects these sub-agents through a predefined execution graph, which may include sequential, parallel, branching, or iterative control flow. This abstraction is particularly useful for complex tasks that require multi-step problem solving rather than a single generation step. For example, a clinical decision-support workflow may retrieve relevant patient information and medical guidelines, invoke specialized agents for diagnosis and verification, and aggregate their outputs into an auditable recommendation. By enforcing structured execution, such workflows improve reliability, reproducibility, and controllability.

These benefits stem from the explicit, program-like execution graphs underlying structured LLM workflows. In this work, we focus on this structured setting, where the control flow and sub-agents are specified before execution. Such explicit graphs enable systematic analysis of workflow-level behavior and expose a well-defined design space for optimization. This scope is distinct from open-ended agentic systems such as ReAct~\citep{yao2022react}, which dynamically interleave reasoning and tool use and may produce substantially different execution traces across queries.

Within this structured setting, optimization is still substantially more challenging than optimizing a single LLM deployment. Each sub-agent can be configured through model selection and reasoning budget, and the workflow structure itself may also expose configurable choices, yielding a combinatorial workflow design space that quickly becomes very large. More importantly, workflow optimization differs from conventional machine learning compilation: instead of optimizing latency while preserving the model's original computation, it must navigate configurations that trade output quality against inference cost. The natural output is therefore not a single fastest implementation, but a set of optimized operating points that support diverse deployment requirements and user preferences.

This frontier-level problem is fundamentally difficult: related formulations of multi-module model assignment are NP-hard~\citep{chen2025llmselector}, and our setting further increases the complexity by expanding the search space to include reasoning budgets and workflow-structure choices. Existing workflow optimization methods~\citep{zhang2025multi,yue-etal-2025-masrouter,su2025difficulty,nie2025resource,chen2025llmselector} largely follow a routing-based paradigm: they learn or tune an inference-time policy to select configurations according to an accuracy--latency objective specified during training. As a result, each policy typically targets a single trade-off point and must be retrained or re-optimized to accommodate different deployment requirements.

Inspired by the machine learning compilers introduced at the beginning, we take a different perspective and argue that structured LLM workflow optimization can be formulated as a compilation problem rather than only as runtime routing. The key distinction is that compilation explores the workflow design space before deployment and produces a reusable set of workflow-level configurations, rather than selecting one configuration online for a particular trade-off objective. We introduce \textbf{FlowCompile}, an optimizing compiler that performs a single compile-time search over the workflow design space and outputs a reusable set of configurations spanning diverse accuracy--latency trade-offs. FlowCompile profiles sub-agents under different model and reasoning-budget choices, composes these sub-agent-level profiles through a workflow-level proxy, and uses the resulting estimates to efficiently explore the workflow configuration space. This compiler-style decomposition avoids exhaustive full-workflow profiling while preserving a flexible set of operating points for deployment. Experiments across diverse workflows and benchmarks show that FlowCompile consistently outperforms heuristically optimized workflows and routing-based baselines by a large margin. We summarize our contributions as follows.

\begin{itemize}

\item We introduce \emph{workflow compilation}, a compiler-inspired paradigm for optimizing structured LLM workflows before deployment and producing reusable accuracy--latency trade-off sets.

\item We develop a structure-aware compositional proxy that lifts reusable sub-agent profiles to workflow-level accuracy and latency estimates, enabling scalable design-space exploration.

\item We present FlowCompile, an optimizing compiler that performs a single compile-time search over model choices, reasoning budgets, and workflow structures, consistently improving accuracy--latency trade-offs across diverse workflows and benchmarks.

\end{itemize}
\section{Related Work}
\label{sec:related_work}

\noindent\textbf{Structured LLM Workflow Optimization.} Structured LLM workflows coordinate multiple LLM-based sub-agents under a predefined execution graph, but often incur substantial latency and inference overhead. Existing efficiency-oriented methods predominantly follow a routing-based paradigm. Representative methods include MaAS~\citep{zhang2025multi}, MasRouter~\citep{yue-etal-2025-masrouter}, and DAAO~\citep{su2025difficulty}, which make inference-time decisions over models and collaboration strategies. Similarly, \citet{nie2025resource} rewrites a workflow into a fixed program and learns an online policy to allocate backends to its components under streaming feedback. \textsc{LLMSELECTOR}~\citep{chen2025llmselector} is closely related because it also leverages module-level assessments to optimize multi-module workflows, but it selects a single static configuration that maximizes accuracy without explicitly modeling cost or latency. DSPy~\citep{khattab2023dspy} also frames LM pipeline optimization as compilation, but it mainly optimizes prompts and demonstrations for improving pipeline accuracy, rather than workflow-level execution trade-offs. 

Our work is complementary to these approaches but addresses a different compiler-inspired formulation: instead of optimizing prompts or learning an inference-time routing policy, FlowCompile performs compile-time workflow-level design-space exploration and produces a reusable set of configurations spanning accuracy--latency trade-offs, without retraining or online adaptation.

\noindent\textbf{Machine Learning Compilers.} Machine learning compilers optimize high-level computational graphs by decomposing them into lower-level optimization units and searching over implementation choices under hardware-aware cost models. TVM~\citep{chen2018tvm} is a representative end-to-end deep learning compiler that combines graph-level optimizations, such as operator fusion and layout transformation, with operator-level code generation and autotuning. AutoTVM~\citep{chen2018learning} further automates tensor-operator optimization by using learned cost models to guide search over large implementation spaces. Ansor~\citep{zheng2020ansor} extends this idea by automatically constructing search spaces and optimizing multiple subgraphs of a neural network through a task scheduler, providing a particularly relevant example of decomposing a full computation graph into local optimization tasks while targeting end-to-end performance.

FlowCompile draws inspiration from this compiler-style decomposition, but targets a different objective and optimization level. ML compilers typically optimize system-level metrics such as latency or memory while preserving the intended computation and output quality of the model. FlowCompile instead performs workflow-level optimization over structured LLM workflows, where model choices and reasoning budgets jointly affect answer quality and inference efficiency, creating an inherent accuracy--latency trade-off. The desired output is therefore a set of operating points rather than a single optimized implementation. Accordingly, in addition to reporting accuracy and latency, we use scalarization-based metrics from multi-objective optimization, such as expected utility, to evaluate trade-off quality~\citep{hayes2021practical,yang2019generalized}.
\section{FlowCompile}
\label{sec:method}

\subsection{Problem Definition and Overview}
\label{sec:problem-definition}

We first formalize \emph{structured LLM workflow compilation}. A structured LLM workflow consists of LLM-based sub-agents connected by a predefined execution graph that specifies sequential, parallel, conditional, or iterative control flow. We denote a structured LLM workflow by $\mathcal{W}=(\mathcal{A},G)$, where $\mathcal{A}$ is the set of sub-agents and $G$ is the workflow execution graph.

Let $\mathcal{C}$ denote the workflow design space. A workflow configuration $c \in \mathcal{C}$ instantiates the executable choices of the workflow, including sub-agent model assignments, reasoning budgets, and optional structural decisions such as branch or refinement-stage execution. A reasoning budget is the maximum number of generated reasoning tokens allocated to a sub-agent call. Executing $c$ on a labeled validation set $\mathcal{D}_{\mathrm{val}}$ induces a workflow-level performance vector $y(c)=(\mathrm{Acc}(c),\mathrm{Lat}(c))$, where $\mathrm{Acc}(c)$ and $\mathrm{Lat}(c)$ denote task accuracy and end-to-end latency.

Given $\mathcal{W}$, $\mathcal{D}_{\mathrm{val}}$, and $\mathcal{C}$, the goal of workflow compilation is to identify a reusable set of high-quality configurations that spans the workflow's accuracy--latency trade-off space, enabling selection under different inference-time latency budgets or performance preferences. Exhaustively evaluating all configurations on $\mathcal{D}_{\mathrm{val}}$ to construct this trade-off set is infeasible due to the combinatorial design space: with $V$ sub-agents, $M$ model choices, and $B$ reasoning-budget options, model--budget assignment alone yields $(MB)^V$ configurations, before structural choices. Even a five-sub-agent workflow with five models and four budgets gives 3.2M configurations, making exhaustive evaluation impractical.

\begin{figure}[t]
    \centering
    \includegraphics[width=\linewidth]{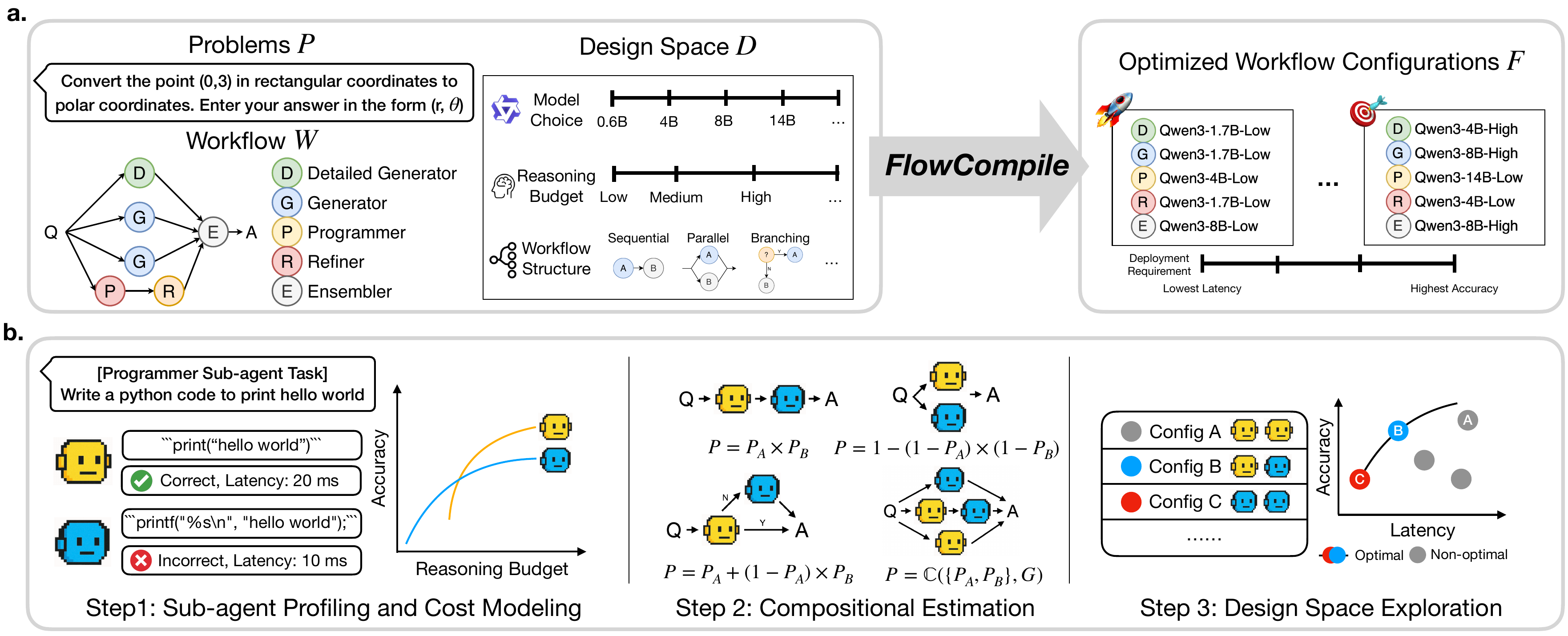}
    \caption{\textbf{Overview of FlowCompile.}
(a) FlowCompile treats structured LLM workflow optimization as compilation: given a problem set, an input workflow, and a design space, it outputs a compiled set of optimized configurations spanning low-latency to high-accuracy deployment regimes.
(b) FlowCompile compiles the workflow through three stages: sub-agent profiling and cost modeling, structure-aware compositional estimation of workflow-level accuracy and latency, and design-space exploration to identify configurations spanning the accuracy--latency trade-off frontier.}
    \label{fig:method}
\end{figure}

FlowCompile addresses this challenge through a compiler-style pipeline. As shown in Figure~\ref{fig:method}, it compiles a structured LLM workflow in three stages: sub-agent profiling, workflow-level estimation, and design-space exploration. Given a workflow specification and a labeled validation set, FlowCompile constructs reusable sub-agent profiles, composes them through lightweight workflow-level estimation, and searches the resulting design space to produce optimized configurations spanning diverse accuracy--latency trade-offs. We describe these stages below.

\subsection{Sub-Agent Data Induction and Profiling}
\label{sec:sub-agent-profile}

FlowCompile first constructs component-level cost models for each sub-agent. Since supervision in $\mathcal{D}_{\mathrm{val}}$ is available only for final workflow outputs, FlowCompile induces sub-agent-level datasets from workflow traces. It executes the workflow on $\mathcal{D}_{\mathrm{val}}$ using a high-capacity reference model, such as GPT-5~\citep{singh2025openai}, records intermediate inputs and outputs for each sub-agent call, and applies an LLM-as-a-judge filter to retain calls that are well-executed and contribute to a correct final answer. The judge prompt is provided in Appendix~\ref{appendix:judge_prompt_data_generation}. For each sub-agent $a\in\mathcal{A}$, the retained examples form an induced dataset $\mathcal{D}_a$, which serves as pseudo ground truth for profiling.

For each sub-agent $a$, we define a discrete sub-agent configuration space $\mathcal{Q}_a=\mathcal{M}_a\times\mathcal{R}_a$, where $\mathcal{M}_a$ is the set of candidate models and $\mathcal{R}_a$ is the set of reasoning budgets. For each configuration $q=(m,r)\in\mathcal{Q}_a$, FlowCompile evaluates sub-agent $a$ on $\mathcal{D}_a$ and records empirical accuracy and latency:
$
\phi_a(q)
=
f_{\mathrm{profile}}(a, q; \mathcal{D}_a)
=
\bigl(\hat{p}_a(q), \hat{\ell}_a(q)\bigr),
$
where $\hat{p}_a(q)$ denotes the profiled sub-agent accuracy and $\hat{\ell}_a(q)$ denotes the profiled sub-agent latency. The accuracy $\hat{p}_a(q)$ is computed against the pseudo ground truth in $\mathcal{D}_a$, using task-specific matching when available and an LLM-as-a-judge otherwise; details of the profiling evaluation protocol are provided in Appendix~\ref{appendix:judge_prompt_profiling}. The collection of profiles $\{\phi_a(q): a\in\mathcal{A}, q\in\mathcal{Q}_a\}$ forms the component-level cost model used by the compiler and can be reused across workflow-level configurations during design space exploration.

\subsection{Workflow-Level Performance Proxy}
\label{sec:workflow-proxy}

Given sub-agent profiles, FlowCompile estimates workflow-level performance with a workflow-level proxy. Directly obtaining the true performance $y(c)=(\mathrm{Acc}(c),\mathrm{Lat}(c))$ for every configuration $c\in\mathcal{C}$ would require full workflow execution and is infeasible at scale. Instead, FlowCompile estimates $\hat y(c)=(\widehat{\mathrm{Acc}}(c),\widehat{\mathrm{Lat}}(c))$ from reusable sub-agent profiles.

For a configuration $c$, let $G_c$ denote the instantiated workflow graph and $q_a(c)$ the configuration assigned to sub-agent $a$. The proxy is defined as
$
\hat{y}(c)
=
\mathcal{M}_{\theta}
\left(
\{\phi_a(q_a(c))\}_{a \in \mathcal{A}_c},
G_c,
E
\right),
$
where $E$ denotes the deployment execution model, such as an edge deployment setting where LLM calls are queued and executed sequentially. The mapping $\mathcal{M}_\theta$ composes sub-agent-level profiles according to the workflow structure $G_c$ and execution model $E$ to produce workflow-level accuracy and latency estimates. Not all such mappings are suitable: to support reliable configuration search, the proxy must preserve key structural properties of the accuracy--latency space, as formalized below.

\noindent\textbf{Proxy requirement.}
FlowCompile does not require $\mathcal{M}_\theta$ to \textit{exactly} predict absolute performance values. Instead, it relies on the proxy to preserve the relative ordering and dominance structure of configurations in the accuracy--latency space, so that high-quality configurations can be identified during search. We formalize this requirement through the following properties.

\textbf{Assumption 1 (Frontier consistency).}
Configurations that are non-dominated under the true performance are likely to remain non-dominated under the proxy estimates, while strongly dominated configurations are unlikely to be identified as part of the estimated frontier.

\textbf{Assumption 2 (Local order preservation).}
For configurations near the trade-off frontier, the proxy approximately preserves relative performance ordering: if $y(c_i) \succeq y(c_j)$, then $\hat{y}(c_i) \succeq \hat{y}(c_j)$ holds with high probability.

These two properties capture the minimal requirements for reliable proxy-based search. Assumption~1 ensures that the non-dominated region of the accuracy--latency space is not substantially distorted by the proxy, so that the identified configuration set remains high-quality. Assumption~2 further ensures that the local ranking among high-quality configurations is preserved, enabling accurate selection under different latency budgets or performance preferences. We next describe a concrete proxy instantiation designed to satisfy these requirements in practice.

\noindent\textbf{Proxy instantiation.}
The proxy $\mathcal{M}_{\theta}$ can be instantiated using analytical rules, learned estimators, or hybrid models. We adopt a simple structure-aware analytical proxy that is lightweight, training-free, and generalizes across workflow structures and deployment settings. This instantiation is designed to preserve ordering and dominance relationships while remaining computationally efficient.

\noindent\textbf{Accuracy proxy.}
Let $\hat{p}_a$ denote the profiled accuracy of sub-agent $a$. FlowCompile composes these values according to workflow control-flow semantics to obtain a structure-aware estimate of workflow-level accuracy. Here, ``sequential'' and ``parallel'' describe the logical structure of the workflow graph, rather than the physical execution schedule of LLM calls, batching, or hardware parallelism. Bounded loops are handled by unrolling them into the corresponding sequence of conditional workflow stages:
\begin{equation}
\begin{aligned}
\hat{p}_{\mathrm{seq}} &= \prod_{i=1}^{N}\hat{p}_{a_i}
&& \text{(logical sequential composition)},\\
\hat{p}_{\mathrm{or}} &= 1-\prod_{i=1}^{N}(1-\hat{p}_{a_i})
&& \text{(disjunctive parallel branches)},\\
\hat{p}_{\mathrm{and}} &= \prod_{i=1}^{N}\hat{p}_{a_i}
&& \text{(conjunctive parallel branches)},\\
\hat{p}_{\mathrm{cond}} &= \hat{p}_{a_1}+(1-\hat{p}_{a_1})\hat{p}_{a_2}
&& \text{(conditional composition)}.
\end{aligned}
\end{equation}
Together, these rules define the recursive estimator:
$
\widehat{\mathrm{Acc}}(c)=\mathcal{C}_{\mathrm{acc}}(\{\hat{p}_a(q_a(c))\},G_c)
$.
This formulation serves as a structure-aware proxy rather than a full probabilistic model, prioritizing efficient and scalable configuration search over exact modeling of sub-agent interactions.

\noindent\textbf{Latency proxy.}
Let $\hat{\ell}_a$ denote the profiled latency of sub-agent $a$. We estimate workflow-level latency with an expected-latency rule, $\widehat{\mathrm{Lat}}(c)=\mathcal{C}_{\mathrm{lat}}(\{\hat{\ell}_a(q_a(c))\}_{a\in A_c},G_c,E)$, where $E$ is the deployment execution model. Under our edge execution model, LLM calls run sequentially, so unconditional stages are summed. Conditional branches are weighted by execution probabilities; e.g., if $a_2$ runs only when $a_1$ fails, then $\widehat{\mathrm{Lat}}_{\mathrm{cond}}=\hat{\ell}_{a_1}+(1-\hat{p}_{a_1})\hat{\ell}_{a_2}$. Bounded retry loops are unrolled and composed similarly. Other execution models, such as critical-path latency under parallel execution, can be handled by replacing $\mathcal{C}_{\mathrm{lat}}$ without re-profiling sub-agents.

Workflow-specific proxy instantiations are provided in Appendix~\ref{appendix:workflow_proxy_details}.
Section~\ref{sec:proxy-validation} empirically validates the proxy assumptions, showing that lightweight composition over reusable sub-agent profiles reliably identifies high-quality configurations without costly end-to-end workflow execution.

\subsection{Design Space Exploration and Deployment}
\label{sec:tradeoff-search}

\noindent\textbf{Trade-off set construction.}
Given the workflow-level proxy, FlowCompile performs lightweight compile-time exploration in two steps.
First, it applies sub-agent-level pruning: for each sub-agent, configuration $q_1$ is removed if it is dominated by $q_2$, i.e., $\hat{p}_a(q_2)\geq\hat{p}_a(q_1)$ and $\hat{\ell}_a(q_2)\leq\hat{\ell}_a(q_1)$, with at least one strict inequality.
Under a monotone workflow-level proxy, this pruning preserves non-dominated workflow configurations while reducing the search space.
FlowCompile then enumerates the remaining configurations $\widetilde{\mathcal{C}}$, computes $\hat{y}(c)=(\widehat{\mathrm{Acc}}(c),\widehat{\mathrm{Lat}}(c))$, and applies non-dominated sorting~\citep{kung1975finding} to obtain the proxy-estimated trade-off set $\widehat{\mathcal{F}}$.

\noindent\textbf{Using the compiled set.}
Once $\widehat{\mathcal{F}}$ is obtained, deployment no longer requires searching the full combinatorial design space; instead, it reduces to lightweight selection among compiled configurations.
We consider three usage settings: latency-constrained deployment, which selects the most accurate configuration satisfying a latency budget; preference-based deployment, which selects the configuration maximizing expected utility under a given accuracy--latency preference; and routing-based adaptation, which uses $\widehat{\mathcal{F}}$ as a compact candidate pool for per-query routing.
The first two are evaluated in Section~\ref{sec:compiled-quality}, and the third in Section~\ref{sec:additional-analyses}.

\noindent\textbf{Compilation cost.}
FlowCompile incurs no model-training cost.
Its main cost is sub-agent profiling, which scales as $\sum_{a\in\mathcal{A}}|\mathcal{Q}_a|$ for a fixed profiling set rather than as the full workflow design space $|\mathcal{C}|$.
The remaining workflow-level estimation and trade-off set construction are lightweight numerical steps over cached profiles and can be completed quickly on a CPU.
After compilation, runtime deployment selects from the compiled configuration set and does not rerun the compilation process.
Appendix~\ref{appendix:cost} provides detailed cost analysis, comparison with exhaustive workflow evaluation, and scalability discussion.
\section{Experiments}
\label{sec:exp}

\subsection{Settings}
\label{sec:exp_settings}

\noindent\textbf{Metric.}
Our evaluation metrics include task accuracy and end-to-end latency. Latency is measured on a single H100 GPU with vLLM~\citep{kwon2023efficient} as the inference engine. For preference-aware evaluation, we follow the standard utility-based view in multi-objective decision making~\citep{hayes2021practical} and use expected utility as the scalar evaluation metric. In our two-objective (accuracy and latency) setting, the utility weight reduces to a single preference parameter $\alpha$: smaller $\alpha$ prioritizes latency efficiency, while larger $\alpha$ prioritizes accuracy. To assess proxy estimation quality and ranking consistency, we report Spearman correlation $\rho$, pairwise agreement, and calibrated mean absolute error (cMAE). Formal definitions are provided in Appendix~\ref{appendix:metric_desc}.

\noindent\textbf{Design Space.}
FlowCompile searches a unified configuration space spanning three dimensions:
(1) \emph{Model size}, using the Qwen-3 family~\citep{yang2025qwen3} with sizes {0.6B, 1.7B, 4B, 8B, 14B};
(2) \emph{Reasoning budget}, with discrete budgets ranging from 10 to 16{,}000 tokens, enforced via budget forcing~\citep{muennighoff2025s1};
and (3) \emph{Workflow structure}, based on AFlow workflows~\citep{zhang2024aflow}, where each sub-agent can be optionally executed, except for \texttt{SelfEnsemble}, which is required to aggregate multiple branches. Additional details are provided in Appendix~\ref{appendix:exp_details}.

\noindent\textbf{Datasets.} We evaluate on four public benchmarks spanning three domains: (i) mathematical reasoning, including \textbf{GSM8K}~\citep{cobbe2021training} and \textbf{MATH-500}~\citep{lightman2023let}; (ii) multi-hop question answering, including \textbf{HotpotQA}~\citep{yang2018hotpotqa}; and (iii) code reasoning, including \textbf{LiveCodeBench}~\citep{jain2024livecodebench}. We construct disjoint profiling and evaluation subsets: the profiling subset is used for sub-agent profiling and baseline training, while all reported results are measured on the held-out evaluation subset. Detailed split protocols are provided in Appendix~\ref{appendix:dataset_splits}.

\noindent\textbf{Baselines.}
We compare FlowCompile against three categories of baselines:
(1) \emph{Single-model baselines}, which directly apply a single large reasoning model without workflows, including Qwen3-32B~\citep{yang2025qwen3} and QwQ-32B~\citep{qwq32b};
(2) \emph{Fixed-workflow baselines}, which execute the same AFlow workflow with a fixed model assignment for all sub-agents;
and (3) \emph{Routing-based baselines}, which adapt model assignment or workflow structure at runtime using query-dependent routers. These include a KNN-based model router implemented with the LLM Router framework~\citep{llmrouter2025}, and MaAS~\citep{zhang2025multi}, which performs dynamic structural routing. For preference-aware evaluation, we include two more baselines that condition on the target accuracy--latency preference. \emph{Pref-Aware Router} performs model-size routing within a fixed workflow, while \emph{Pref-Aware MaAS} combines MaAS-style structural routing with model-size selection; both favor smaller models when latency is prioritized over accuracy. All workflow baselines use the same AFlow structures and Qwen-3 model family as our method for fairness.

\begin{figure*}[t]
    \centering
    \begin{minipage}[t]{0.42\textwidth}
        \vspace{0pt}
        \centering
        \includegraphics[width=\linewidth]{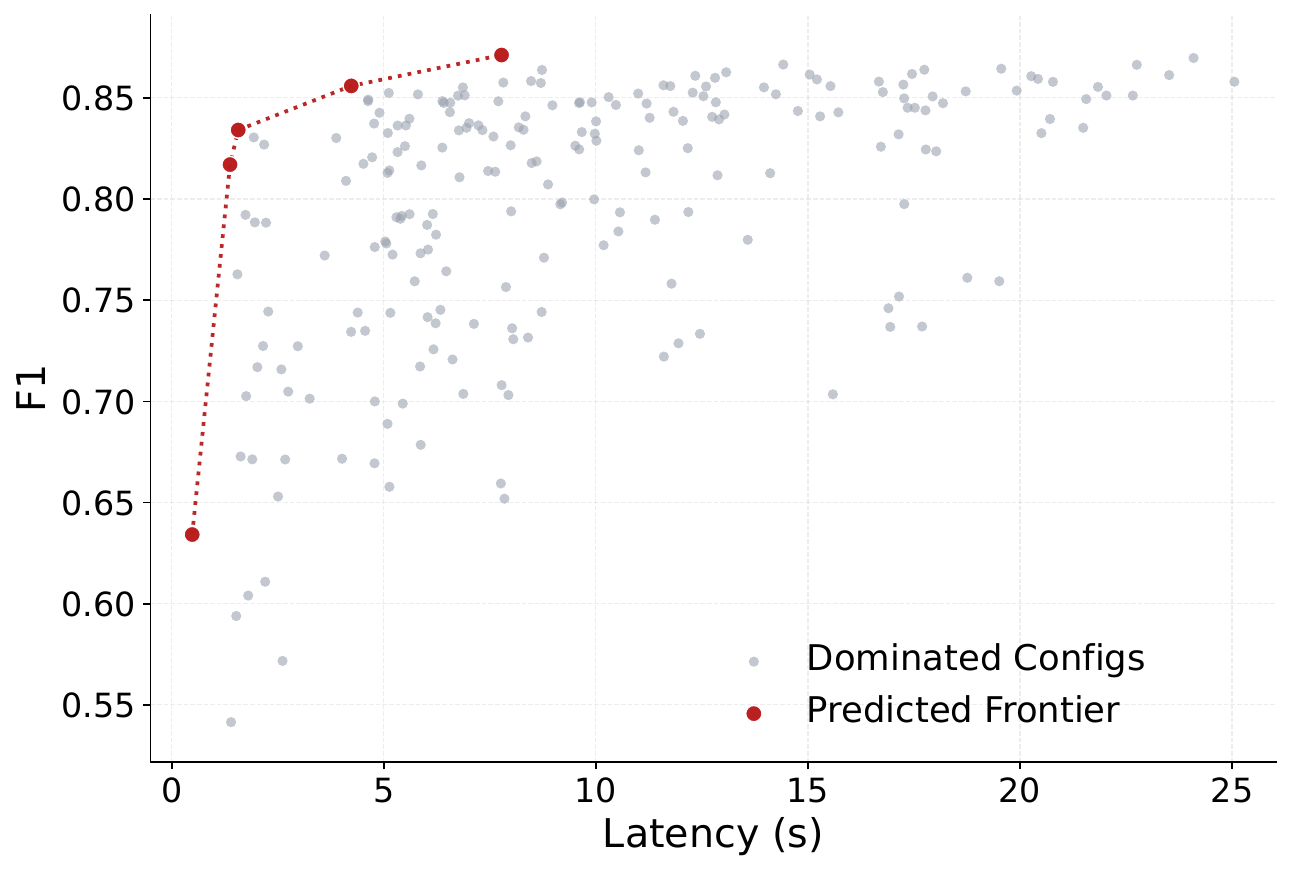}
        \captionof{figure}{\textbf{Frontier consistency.} The proxy-estimated frontier matches the empirically measured frontier on HotpotQA under the restricted space.}
        \label{fig:bruteforce_exp}
    \end{minipage}
    \hfill
    \begin{minipage}[t]{0.55\textwidth}
        \vspace{0pt}
        \centering
        \captionof{table}{\textbf{Local order preservation within compiled configurations.} The proxy preserves the ordering of high-quality configurations and yields low absolute error after systematic bias correction.}
        \label{tab:estimation_accuracy}
        \vspace{-0.5em}

        \scriptsize
        \setlength{\tabcolsep}{4pt}
        \renewcommand{\arraystretch}{0.92}
        \begin{tabular}{lccc}
        \toprule
        \multicolumn{4}{c}{\textbf{Latency Estimation}} \\
        \textbf{Benchmark}
        & Spearman ($\uparrow$)
        & Pairwise ($\uparrow$)
        & cMAE ($\downarrow$) \\
        \midrule
        GSM8K & 0.98 & 0.96 & 2.8 \\
        MATH-500 & 0.97 & 0.96 & 10.8 \\
        HotpotQA & 0.99 & 0.98 & 0.1 \\
        LiveCodeBench & 0.90 & 0.89 & 3.8 \\
        \textbf{Average} & \textbf{0.96} & \textbf{0.95} & \textbf{4.4} \\
        \midrule
        \multicolumn{4}{c}{\textbf{Accuracy Estimation}} \\
        \textbf{Benchmark}
        & Spearman ($\uparrow$)
        & Pairwise ($\uparrow$)
        & cMAE ($\downarrow$) \\
        \midrule
        GSM8K & 0.93 & 0.89 & 0.015 \\
        MATH-500 & 0.81 & 0.81 & 0.047 \\
        HotpotQA & 0.98 & 0.95 & 0.013 \\
        LiveCodeBench & 0.97 & 0.95 & 0.015 \\
        \textbf{Average} & \textbf{0.92} & \textbf{0.90} & \textbf{0.023} \\
        \bottomrule
        \end{tabular}
    \end{minipage}
\end{figure*}

\subsection{Proxy Validation}
\label{sec:proxy-validation}

FlowCompile relies on a workflow-level proxy to estimate configuration accuracy and latency. Since the proxy is an approximation rather than an exact simulator, we empirically validate whether it satisfies the two assumptions from Section~\ref{sec:workflow-proxy} that are required for reliable configuration search.

\noindent\textbf{Frontier consistency.} We first test whether the proxy can recover the empirical accuracy--latency frontier by exhaustively evaluating all workflows in a restricted design space where brute-force evaluation is feasible. This restricted design space preserves the essential structure of the full design space while using a compact set of sub-agent configurations. We use HotpotQA and LiveCodeBench, which cover both simpler and more complex workflows as well as the proxy composition rules; details are provided in Appendix~\ref{appendix:frontier_consistency_bruteforce}. Figures~\ref{fig:bruteforce_exp} and~\ref{fig:livecodebench_bruteforce} show that the proxy-estimated configurations (red dots) closely align with the empirically measured non-dominated region, while most remaining configurations are dominated in the measured accuracy--latency space. This supports the frontier-consistency assumption and shows that the proxy can guide search toward high-quality configurations without exhaustive end-to-end evaluation.

\noindent\textbf{Local order preservation.} We next test whether the proxy preserves the ordering of high-quality configurations in the full design space. For each benchmark, we sample 20 configurations from the compiled set, run full test-set evaluation, and compare proxy-estimated and measured accuracy/latency using the metrics in Section~\ref{sec:exp_settings}. Table~\ref{tab:estimation_accuracy} shows that the proxy largely preserves the ordering of high-quality configurations, with average pairwise agreement of 0.90 for accuracy and 0.95 for latency. MATH-500 is the most challenging case for accuracy estimation, but the proxy still preserves most pairwise orderings. After correcting systematic bias, the proxy achieves low absolute error, with an average latency error of 4.4 seconds and an average accuracy error of 2.3 percentage points. These results support the local order-preservation assumption, enabling reliable preference-aware selection.

Together, these results show that, although the proxy is not assumed to be exact, it preserves the key trade-off structure needed to guide FlowCompile's configuration search.

\begin{table}[t]
\centering
\small
\caption{\textbf{Preference-aware evaluation under heterogeneous preferences.}
Expected utility across four benchmarks under randomly sampled per-query preferences, reported as mean $\pm$ std.}
\label{tab:main_results}
\begin{tabular}{@{}lccccl@{}}
\toprule
 & \multicolumn{4}{c}{\textbf{Benchmarks}} & \multicolumn{1}{c}{} \\
\multirow{-2}{*}{\textbf{Method}} & GSM8K & MATH-500 & HotpotQA & LiveCodeBench & \multicolumn{1}{c}{\multirow{-2}{*}{\textbf{Avg.}}} \\ \midrule
\multicolumn{5}{c}{\textit{Single Model}} &  \\
Qwen3-32B        & $74.1\pm0.6$  & $82.2\pm0.3$ & $67.4\pm0.4$  & $55.9\pm0.3$ & 69.9 \\
QwQ-32B          & $61.6\pm1.2$  & $80.6\pm0.4$ & $62.6\pm0.7$  & $54.6\pm0.4$ & 64.9 \\
\midrule
\multicolumn{5}{c}{\textit{Fixed Workflow}} &  \\
Qwen3-4B    & $69.9\pm0.7$ & $70.3\pm0.7$ & $68.3\pm0.4$ & $61.0\pm0.3$ & 67.4 \\
Qwen3-8B    & $57.6\pm0.9$ & $48.5\pm1.4$ & $58.1\pm0.8$ & $49.1\pm0.6$ & 53.3  \\ \midrule
\multicolumn{5}{c}{\textit{Workflow with Router}} &  \\
MaAS (Qwen3-4B)       & $87.2\pm0.1$ & $78.4\pm0.9$  & $74.2\pm0.5$ & $66.0\pm0.6$  & 76.5 \\
MaAS (Qwen3-8B)       & $82.2\pm0.4$ & $71.9\pm0.6$  & $65.2\pm0.5$ & $55.7\pm0.5$  & 68.8  \\
KNN Router            & $81.9\pm0.2$ & $77.6\pm0.5$  & $74.1\pm0.4$ & $53.6\pm0.3$  & 71.8 \\
Pref-Aware Router     & $68.0\pm0.7$ & $70.1\pm1.1$  & $69.3\pm0.4$ & $59.7\pm0.6$  & 66.8  \\
Pref-Aware MaAS       & $80.7\pm0.2$ & $78.2\pm1.3$  & $78.7\pm0.4$ & $65.5\pm0.8$  & 75.8 \\ \midrule
\textbf{FlowCompile}  & $\boldsymbol{89.2}\pm0.6$ & $\boldsymbol{84.2}\pm0.8$  & $\boldsymbol{88.4}\pm0.4$ & $\boldsymbol{80.1}\pm0.6$ & \textbf{85.5} \\
\bottomrule
\end{tabular}
\end{table}

\begin{figure}[t]
    \centering

    \begin{minipage}[t]{0.48\textwidth}
        \vspace{0pt}
        \centering
        \includegraphics[width=0.85\linewidth]{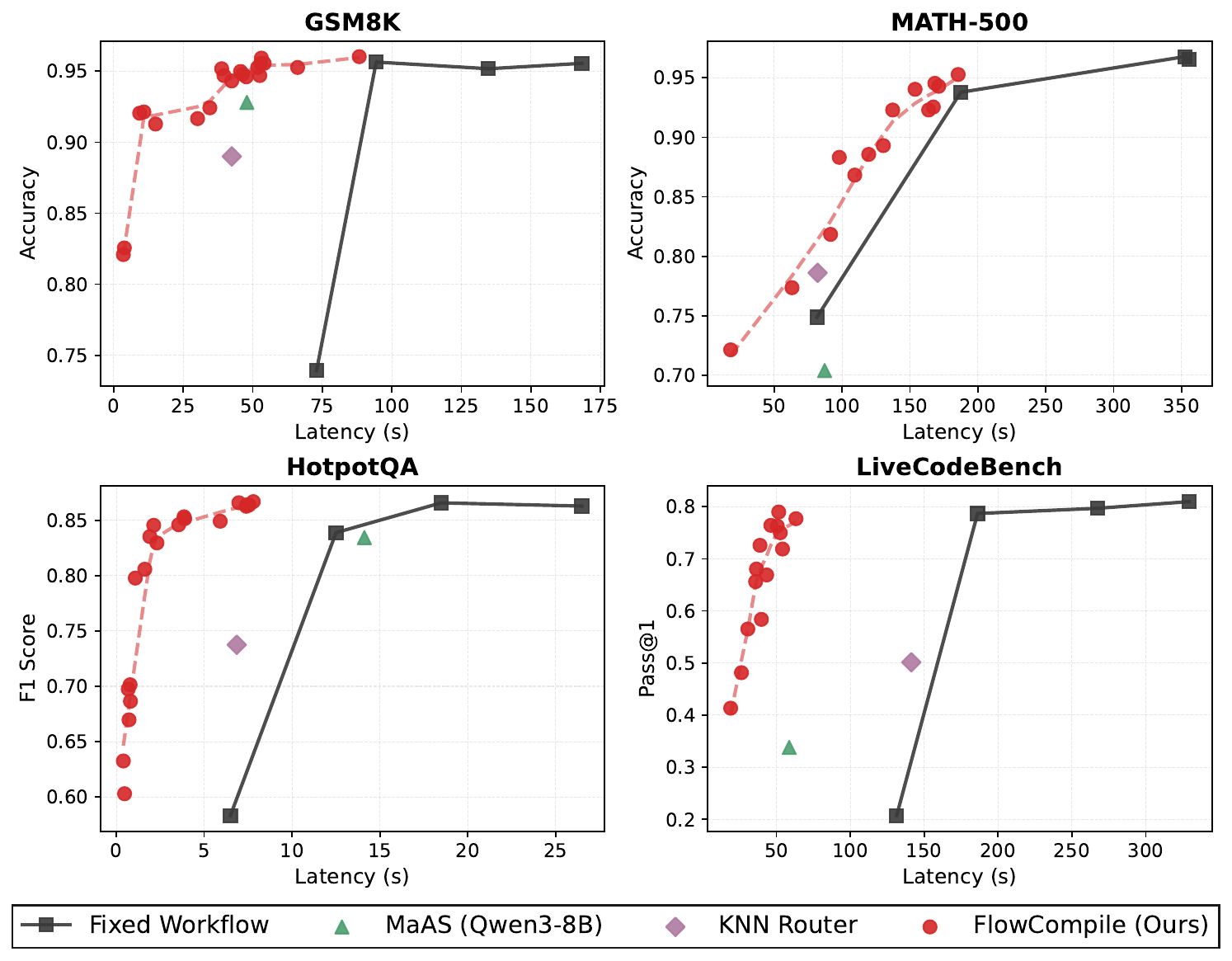}
        \captionof{figure}{\textbf{Empirical accuracy--latency trade-offs.} FlowCompile produces configurations with substantially better measured accuracy--latency trade-offs than baselines.}
        \label{fig:pareto_validity}
    \end{minipage}
    \hfill
    \begin{minipage}[t]{0.48\textwidth}
        \vspace{0pt}
        \centering
        \includegraphics[width=0.9\linewidth]{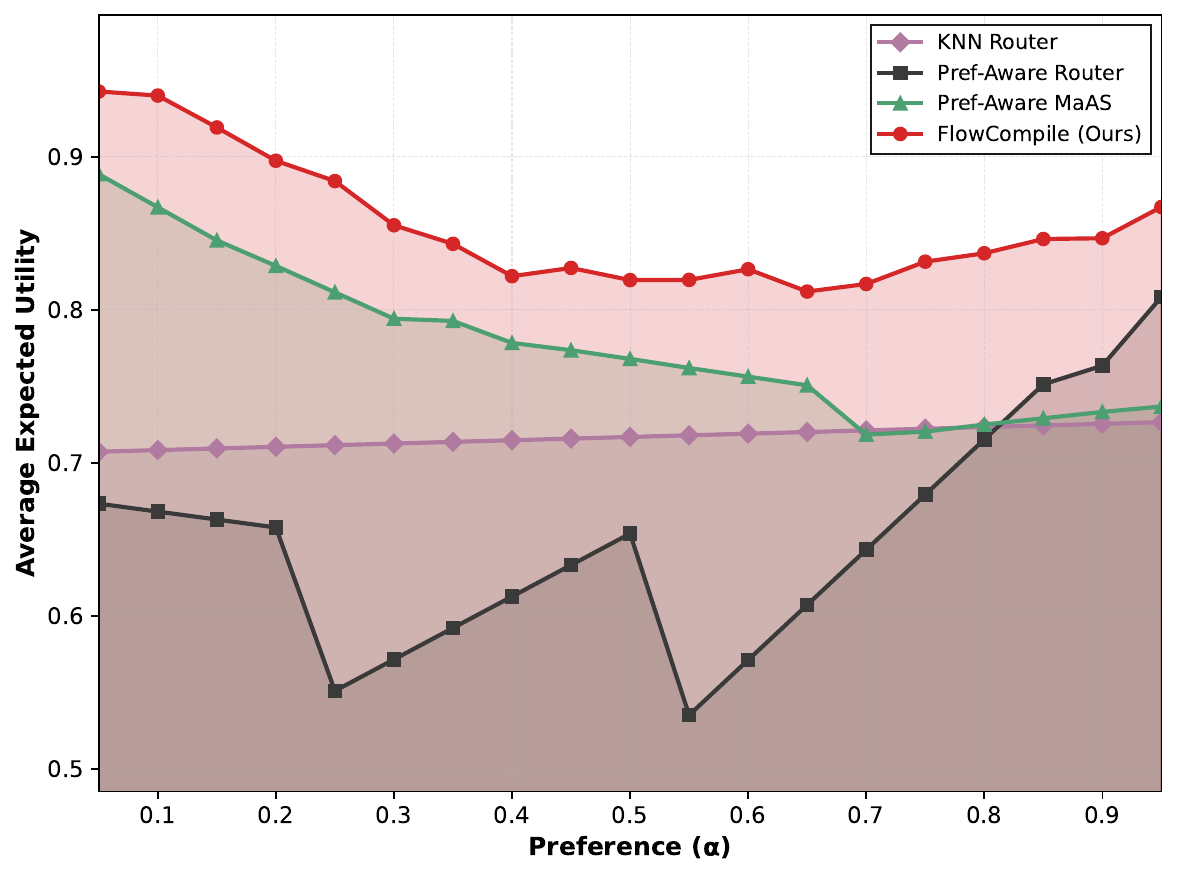}
        \captionof{figure}{\textbf{Preference-aware evaluation under fixed preferences.} FlowCompile achieves the highest expected utility averaged across four benchmarks as the preference varies.}
        \label{fig:average_alpha_sweep}
    \end{minipage}
    \vspace{-2mm}
\end{figure}

\subsection{End-to-End Quality of Compiled Configurations}
\label{sec:compiled-quality}

Having validated the proxy, we evaluate the compiled configurations on the test set by comparing their measured accuracy--latency trade-offs with baselines and assessing their preference-aware performance using expected utility.

\noindent\textbf{Accuracy--latency trade-offs.}
We evaluate the compiled configurations and all baselines under the same inference setting across four benchmarks. As shown in Figure~\ref{fig:pareto_validity}, FlowCompile consistently achieves lower latency at comparable or higher accuracy than the baselines, yielding substantially better accuracy--latency trade-off curves. This comes from jointly optimizing model choice, reasoning budget, and workflow structure, allowing FlowCompile to discover efficient configurations that are difficult to obtain from fixed workflows or runtime routing alone. We also provide numerical accuracy and latency results in Table~\ref{tab:accuracy_latency_tradeoff} in Appendix~\ref{appendix:accuracy_latency_tradeoff}. Since FlowCompile outputs a set of configurations rather than a single operating point, we report two representative selections: an accuracy-priority configuration that aims to match the full Qwen3-14B workflow (marked as full baseline), and a latency-priority configuration that minimizes latency while preserving strong accuracy. The accuracy-priority setting achieves nearly the same accuracy as the full baseline with an average $3.4\times$ speedup and an even greater speedup of $\mathbf{6.4}\times$ in LiveCodeBench, while the latency-priority setting achieves an average $\mathbf{12.7}\times$ speedup with competitive accuracy.

\noindent\textbf{Preference-aware evaluation.}
Representative points and frontier plots characterize accuracy--latency trade-offs, but they do not provide a single comparable measure of how well a method supports different deployment preferences.
We therefore use expected utility, a standard utility-based formulation of multi-objective decision making, as a scalarized measure that combines accuracy and latency efficiency under a specified preference weight and enables direct comparison across methods~\citep{hayes2021practical}.
We vary the preference parameter $\alpha\in(0,1)$ under two settings.
In the heterogeneous setting, each query is assigned a randomly sampled $\alpha$ to represent mixed-preference workloads; we repeat this ten times and report the mean and standard deviation.
In the fixed setting, all queries share the same $\alpha$, and we sweep $\alpha$ across the preference spectrum to evaluate robustness under different global preferences.
Detailed protocols are provided in Appendix~\ref{appendix:per_benchmark_eu}.

Table~\ref{tab:main_results} shows that under the heterogeneous setting, FlowCompile achieves the highest expected utility across all benchmarks, outperforming the strongest baseline by an average of $+7.9$. Figure~\ref{fig:average_alpha_sweep} further shows that, in the fixed setting, FlowCompile achieves the highest average utility across the four benchmarks for every preference value. In contrast, preference-aware routing baselines are effective only over narrower preference ranges. Per-benchmark results in Figure~\ref{fig:per_benchmark_fixed_pref_results} of Appendix~\ref{appendix:per_benchmark_eu} show similar patterns. These results indicate that a single compiled set from FlowCompile can support diverse deployment requirements and user preferences.

\noindent\textbf{Qualitative analysis.}
We further inspect the compiled configurations and find that they form interpretable operating regimes: latency-priority configurations favor simpler workflows and lower-cost model-budget choices, while accuracy-priority configurations allocate compute to task-critical stages. Detailed analysis is provided in Appendix~\ref{appendix:analysis-compiled-workflow-configs}.

\begin{table}[t]
\centering
\small

\begin{minipage}[t]{0.48\linewidth}
\centering
\captionof{table}{\textbf{Cross-benchmark transfer.}
Reusing MATH-500 sub-agent profiles for GSM8K preserves proxy quality and good expected utility.}
\label{tab:estimation-transfer}

\scriptsize
\setlength{\tabcolsep}{3pt}
\begin{tabular}{lccc}
\toprule
 & \multicolumn{2}{c}{\textbf{Correlation ($\rho$)}} 
 & \multirow{2}{*}{\textbf{Expected Utility}} \\
 & Latency & Accuracy &  \\ 
\midrule
GSM8K & 0.98 & 0.93 & 88.77 \\
GSM8K (transfer) & 0.94 & 0.76 & 86.43 \\
\bottomrule
\end{tabular}
\end{minipage}
\hfill
\begin{minipage}[t]{0.48\linewidth}
\centering
\captionof{table}{\textbf{Design space ablation.}
Expected utility improves as FlowCompile expands the searchable design space.}
\label{tab:ablation-search-space}

\scriptsize
\setlength{\tabcolsep}{4pt}
\begin{tabular}{lc}
\toprule
\textbf{Design Space} & \textbf{HotpotQA} \\
\midrule
Model & 69.7 \\
Model + Budget & 83.2 \\
Model + Budget + Structure (Ours) & \textbf{88.4} \\
\bottomrule
\end{tabular}
\end{minipage}

\vspace{-2mm}
\end{table}

\subsection{Additional Analysis}
\label{sec:additional-analyses}

We further analyze FlowCompile's behavior and design choices. Unless otherwise specified, all experiments use the heterogeneous preference-aware evaluation setting.

\noindent\textbf{Per-query routing.}
FlowCompile is complementary to routing-based methods because the compiled configuration set provides a compact candidate pool for query-level adaptation. To demonstrate this, we combine FlowCompile with a simple KNN router ($k=20$) on GSM8K and MATH-500, allowing the router to select among compiled configurations for each query. As shown in Table~\ref{tab:per-query-routing} in Appendix~\ref{appendix:per_query_results}, this further improves expected utility over FlowCompile alone, with gains of $+2.6$ and $+6.3$ on GSM8K and MATH-500, respectively. Appendix~\ref{appendix:per_query_results} provides detailed analysis of why routing improves performance: the compiled set offers a high-quality candidate pool, and the router makes query-conditioned selections within this set that correlate with problem difficulty. These results highlight the complementarity between compile-time optimization and runtime routing: FlowCompile constructs a strong trade-off set, while routing performs lightweight query-level selection within it.

\noindent\textbf{Cross-benchmark transfer.}
Since sub-agent profiling is the main cost of FlowCompile, we evaluate whether profiles can be reused across related tasks sharing the same workflow. We reuse sub-agent data profiled on MATH-500 to compile configurations for GSM8K, without re-profiling on the GSM8K validation set, and evaluate the resulting configurations on the GSM8K test set. As shown in Table~\ref{tab:estimation-transfer}, transferred profiles preserve strong latency correlation, reasonable accuracy correlation, and competitive expected utility. This suggests that FlowCompile can reuse profiles across related tasks to reduce profiling cost while still producing competitive compiled configurations.

\noindent\textbf{Ablation study.}
We ablate two design choices on HotpotQA.
First, we ablate the design space by progressively adding model choice, reasoning-budget choice, and workflow-structure choice; expected utility consistently improves with more dimensions (Table~\ref{tab:ablation-search-space}).
Second, we test sensitivity to the reference model used for sub-agent data induction by replacing GPT-5 with GPT-5-mini or Qwen3-1.7B for sub-agent data induction. The resulting frontiers are highly consistent (Figure~\ref{fig:reference_model_ablation}), suggesting robustness as long as the reference model can generate reasonable workflow traces.
\section{Conclusion}
\label{sec:conclusion}

FlowCompile is an optimizing compiler for structured LLM workflows. We argue that structured workflow optimization should be treated as a compilation problem, rather than only as runtime routing. By turning workflow optimization into a reusable compile-time artifact, FlowCompile enables configurations to be selected and adapted under diverse deployment requirements. More broadly, our results motivate \emph{workflow compilation} as a systems abstraction for future LLM applications, where efficiency, controllability, and adaptability must be managed at the workflow level.

\bibliographystyle{plainnat}
\bibliography{main}


\appendix

\newpage

\section{FlowCompile Algorithm}
\label{appendix:algorithm}

\begin{algorithm}[h]
\caption{\textsc{FlowCompile}: Structured LLM Workflow Compilation}
\label{alg:flowcompile}
\KwIn{
Workflow $\mathcal{W}=(\mathcal{A},G)$ with sub-agents $\mathcal{A}$ and structure space; \\
Workflow-level labeled validation set $\mathcal{D}_{\mathrm{val}}$; \\
Reference model $m_{\mathrm{ref}}$; \\
Sub-agent model choices $\{\mathcal{M}_a\}_{a\in\mathcal{A}}$ and reasoning budgets $\{\mathcal{R}_a\}_{a\in\mathcal{A}}$; \\
Deployment execution model $\mathcal{E}$.
}
\KwOut{
Proxy-estimated optimized configuration set $\widehat{\mathcal{F}}$.
}

\BlankLine
\textbf{Step 1: Sub-agent data induction and profiling}

Execute $\mathcal{W}$ on $\mathcal{D}_{\mathrm{val}}$ using $m_{\mathrm{ref}}$ to collect workflow traces\;

Apply LLM-as-a-judge filtering to retain high-quality intermediate input--output pairs\;

Construct induced sub-agent datasets $\{\mathcal{D}_a\}_{a\in\mathcal{A}}$\;

\ForEach{sub-agent $a \in \mathcal{A}$}{
    Define $\mathcal{Q}_a = \mathcal{M}_a \times \mathcal{R}_a$\;
    \ForEach{sub-agent configuration $q=(m,r)\in\mathcal{Q}_a$}{
        Profile $a$ on $\mathcal{D}_a$ under $q$ to obtain
        $\phi_a(q)=\bigl(\hat p_a(q), \hat \ell_a(q)\bigr)$\;
    }
    Remove locally dominated configurations from $\mathcal{Q}_a$\;
}

\BlankLine
\textbf{Step 2: Workflow-level compositional estimation}

Enumerate the pruned workflow configuration space $\widetilde{\mathcal{C}}$\;

\ForEach{workflow configuration $c \in \widetilde{\mathcal{C}}$}{
    Let $G_c$ be the instantiated workflow graph and $q_a(c)$ be the configuration assigned to sub-agent $a$\;
    
    Estimate workflow accuracy
    $\widehat{\mathrm{Acc}}(c)
    = \mathcal{C}_{\mathrm{acc}}\bigl(\{\hat p_a(q_a(c))\}_{a\in\mathcal{A}_c}, G_c\bigr)$\;
    
    Estimate workflow latency
    $\widehat{\mathrm{Lat}}(c)
    = \mathcal{C}_{\mathrm{lat}}\bigl(\{\hat \ell_a(q_a(c))\}_{a\in\mathcal{A}_c}, G_c, \mathcal{E}\bigr)$\;
    
    Set $\hat y(c)=\bigl(\widehat{\mathrm{Acc}}(c),\widehat{\mathrm{Lat}}(c)\bigr)$\;
}

\BlankLine
\textbf{Step 3: Trade-off set construction}

Apply non-dominated sorting to $\{\hat y(c): c\in\widetilde{\mathcal{C}}\}$, treating higher accuracy and lower latency as better\;

Let $\widehat{\mathcal{F}}$ be the resulting proxy-estimated non-dominated configuration set\;

\BlankLine
\Return $\widehat{\mathcal{F}}$
\end{algorithm}

\begin{figure}[t]
    \centering
    \begin{subfigure}[t]{0.75\linewidth}
        \centering
        \includegraphics[width=0.8\linewidth]{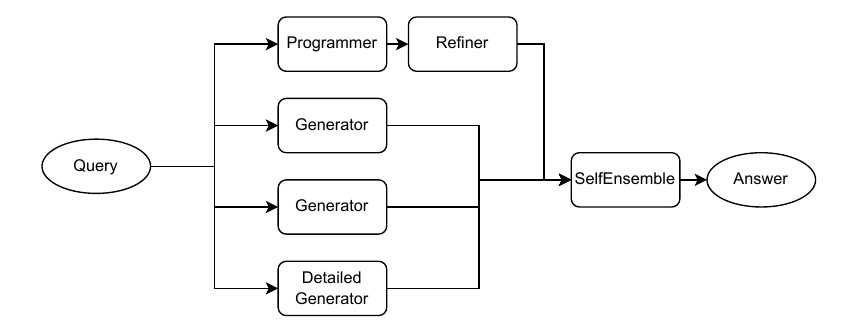}
        \caption{Workflow structure used for GSM8K and MATH-500.}
        \label{fig:workflow_math}
    \end{subfigure}

    \vspace{0.5em}

    \begin{subfigure}[t]{0.75\linewidth}
        \centering
        \includegraphics[width=0.8\linewidth]{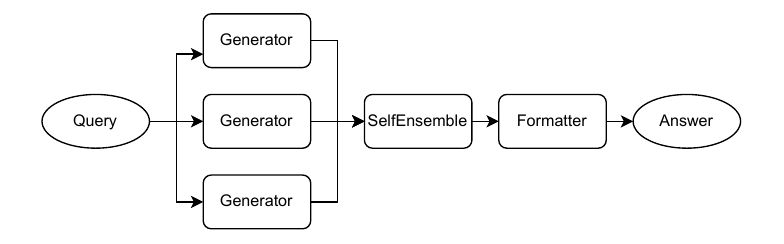}
        \caption{Workflow structure used for HotpotQA.}
        \label{fig:workflow_hotpotqa}
    \end{subfigure}

    \vspace{0.5em}

    \begin{subfigure}[t]{0.75\linewidth}
        \centering
        \includegraphics[width=0.8\linewidth]{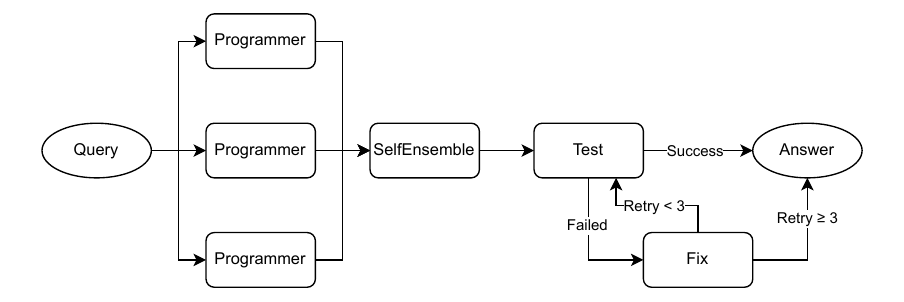}
        \caption{Workflow structure used for LiveCodeBench.}
        \label{fig:workflow_livecodebench}
    \end{subfigure}

    \caption{Workflow structures used by FlowCompile across benchmarks.}
    \label{fig:workflow_structures}
\end{figure}

\section{Additional Experiment Setup Details}
\label{appendix:exp_details}

\subsection{Dataset Splits}
\label{appendix:dataset_splits}

FlowCompile requires a profile set for sub-agent data induction, sub-agent profiling, configuration selection, and baseline training, while final results must be evaluated on held-out examples. We therefore construct disjoint profile and test sets for each benchmark. Unless otherwise specified, we use a 1:4 split, with 20\% of the examples used for profiling and 80\% held out for testing.

For GSM8K, we use the official test set of 1,319 problems and split it into disjoint profile and test sets. For MATH-500, which consists of a single benchmark set of 500 problems, we similarly split the examples into disjoint profile and test sets. For HotpotQA, following AFlow~\citep{zhang2024aflow}, we sample 1,000 examples and split them into disjoint profile and test sets. For LiveCodeBench, we use all released code-generation problems and split them into disjoint profile and test sets with a fixed random seed.

The profile set is used only for compile-time profiling, configuration selection, and baseline training. All reported accuracy, F1, Pass@1, latency, and expected-utility results are measured only on the held-out test set.

\subsection{Latency Measurement and Reasoning Budgets}
\label{appendix:latency_budget_details}

\noindent\textbf{Latency measurement.}
All latency measurements are conducted on a single NVIDIA H100 80GB GPU using the vLLM inference engine. We measure end-to-end model inference latency with a batch size of 1 to ensure accurate measurements in our setting.

\noindent\textbf{Reasoning budget.}
We consider a discrete set of reasoning budgets for each benchmark, chosen to cover a wide range of accuracy--latency trade-offs. For GSM8K, we use [10, 200, 400, 800, 1000, 1500, 2000, 3000, 4000, 5000, 6000, 8000, 10000]. For MATH-500, we use [10, 200, 400, 800, 1000, 1500, 2000, 3000, 4000, 5000, 6000, 7000, 8000, 12000, 16000]. For HotpotQA, we use [10, 50, 100, 200, 300, 400, 500, 600, 700, 800, 900, 1000, 1500, 2000, 4000, 8000]. For LiveCodeBench, we use [10, 200, 400, 800, 1000, 1500, 2000, 3000, 4000, 5000, 6000, 7000, 8000, 12000, 16000].

\subsection{Workflow Structures and Proxy Instantiations}
\label{appendix:workflow_proxy_details}

For each benchmark, we use the full workflow discovered by AFlow~\citep{zhang2024aflow} as the base workflow for FlowCompile (Figure~\ref{fig:workflow_structures}). Importantly, FlowCompile treats the workflow structure itself as part of the configuration space, rather than assuming a fixed execution graph. For GSM8K and MATH-500, we use a shared math workflow in which each of the four branches can be independently activated or skipped. The SelfEnsemble sub-agent is executed only when at least two branches are activated; otherwise, it is skipped. We apply the same SelfEnsemble activation rule to all three workflow types. For HotpotQA, each of the three generator sub-agents is optional, and the formatter can also be skipped. For LiveCodeBench, each of the three programmer sub-agents is optional, and the maximum number of retry attempts is additionally configurable. These structural choices are jointly optimized with model choices and reasoning budgets, and therefore may differ across compiled configurations.

We next provide workflow-specific details for the proxy used in our experiments. All modules, including SelfEnsemble, Formatter, Refiner, and Fix, are treated as profiled sub-agents. Their accuracy and latency are measured on induced sub-agent examples and then composed according to the workflow structure. In particular, SelfEnsemble is not modeled by a closed-form majority-vote rule; FlowCompile directly profiles its aggregation behavior on induced aggregation examples.

\noindent\textbf{GSM8K and MATH-500.}
The math workflow contains four optional solving branches: a Programmer--Refiner branch, two Generator branches, and a Detailed Generator branch. For the Programmer--Refiner branch, the branch-level accuracy is computed by sequential composition, $\hat{p}_{\mathrm{prog}}\hat{p}_{\mathrm{ref}}$. The other solving branches use their profiled accuracies directly. Given the active branches in a configuration, FlowCompile first composes their branch-level accuracies using the disjunctive rule, estimating the probability that at least one active branch produces a correct candidate solution. When at least two branches are active, SelfEnsemble is executed and treated as a profiled aggregation sub-agent; its profiled accuracy is then composed sequentially with the candidate-solution stage. When only one branch is active, SelfEnsemble is skipped and the branch output is used directly. Under the edge execution model, the latency proxy sums the profiled latencies of all active branch calls, including Refiner when the Programmer branch is selected, and adds the profiled SelfEnsemble latency when SelfEnsemble is executed.

\noindent\textbf{HotpotQA.}
The HotpotQA workflow contains up to three optional Generator branches, followed by an optional SelfEnsemble aggregation step and an optional Formatter. The active Generator branches are composed using the disjunctive rule, reflecting that any correct generated answer can support the final prediction. If multiple Generator branches are active, SelfEnsemble is executed as a profiled aggregation sub-agent and composed sequentially with the generator stage. The Formatter, when enabled, is also treated as a profiled sub-agent and composed sequentially with the preceding output. The latency proxy sums the profiled latencies of executed Generator calls and adds the SelfEnsemble and Formatter latencies when those modules are present.

\noindent\textbf{LiveCodeBench.}
The LiveCodeBench workflow contains up to three optional Programmer branches, followed by SelfEnsemble, test execution, and bounded repair through the Fix module. The active Programmer branches are composed using the disjunctive rule, since any correct candidate program can solve the task. SelfEnsemble is treated as a profiled aggregation sub-agent that selects or combines candidate programs before testing. The Test node is not an LLM sub-agent; it deterministically executes the generated code against the public test cases provided by LiveCodeBench and returns a success or failure signal. Since test execution is deterministic and does not involve LLM inference, its latency is excluded from the LLM latency proxy. The repair stage is modeled as a bounded conditional workflow: Fix is executed only when the current program fails the public tests and the retry budget has not been exhausted. Accordingly, the accuracy proxy follows the conditional composition rule over the initial program-generation stage and subsequent repair attempts. The latency proxy uses expected latency: unconditional generation and aggregation stages are summed, while each Fix attempt is weighted by the estimated probability that all previous attempts fail and the retry stage is executed.

\section{Additional Details on Compilation Cost}
\label{appendix:cost}

FlowCompile performs compile-time optimization and does not involve any model training. Its compilation cost consists of three main stages: (1) inducing sub-agent data and profiling sub-agents under candidate configurations, (2) composing workflow-level accuracy and latency estimates from the profiled sub-agent statistics, and (3) constructing the proxy-estimated accuracy--latency trade-off set. Suppose the profiling set contains $D$ data points, the workflow contains $A$ sub-agent roles after unrolling bounded loops, and each sub-agent role has $M$ candidate models and $R$ reasoning-budget options. For simplicity, we describe the cost assuming the same model and budget choices for all sub-agent roles; the expression directly generalizes to heterogeneous sub-agent configuration spaces.

\subsection{Step-by-Step Cost Analysis}
\label{appendix:cost_step_analysis}

\noindent\textbf{Stage 1: Sub-agent data induction and profiling.}
FlowCompile first executes the workflow on the profile set using a reference model to collect workflow traces and induce sub-agent-level profiling data. Bounded iterative workflows are unrolled before profiling, so repeated calls at different retry depths are treated as distinct sub-agent roles, e.g., $\mathrm{Fix}_1$, $\mathrm{Fix}_2$, and $\mathrm{Fix}_3$ in LiveCodeBench. This requires $D$ reference workflow executions. The resulting traces provide induced input--output examples for each executed sub-agent role; because some branches may be skipped, each role contributes up to $D$ examples, yielding at most $D \times A$ examples after unrolling.

FlowCompile then profiles each sub-agent role independently under candidate model and reasoning-budget configurations. This requires up to $D \times A \times M \times R$ sub-agent inference calls. Equivalently, for a fixed profiling set, the number of profiled sub-agent configurations scales as $A M R$, or more generally as $\sum_{a\in\mathcal{A}}|\mathcal{Q}_a|$ for heterogeneous configuration spaces. These calls are independent single-model inferences and do not require executing the full workflow control logic, making this stage highly parallelizable. In our implementation, we use vLLM with large batch sizes to execute this stage efficiently. For example, in our experiments, we use $32$ H100 GPUs with a total batch size of $256$, yielding a typical profiling time of about \textbf{1 hour} per benchmark.

\noindent\textbf{Stage 2: Workflow-level compositional estimation.}
Given the profiled sub-agent statistics, FlowCompile estimates workflow-level accuracy and latency by applying the structure-aware proxy to candidate workflow configurations. This stage involves no model inference or end-to-end workflow execution; it only performs numerical composition over cached sub-agent profiles.

In the worst case, if each sub-agent can be independently activated or skipped and each active sub-agent has $MR$ model--budget choices, the number of candidate workflow configurations scales as $O((1+MR)^A)$. In practice, FlowCompile first applies sub-agent-level Pareto pre-filtering, which removes locally dominated model--budget choices and effectively replaces $MR$ with a much smaller pruned set size. With vectorized numerical operations, the resulting composition step completes in under one second on a CPU.

\noindent\textbf{Stage 3: Trade-off set construction.}
Finally, FlowCompile identifies the proxy-estimated non-dominated configurations from the composed candidates. Let $N$ denote the number of remaining workflow configurations after sub-agent-level pruning and structural enumeration. We apply standard non-dominated sorting using an $O(N \log N)$ algorithm~\citep{kung1975finding}, treating higher accuracy and lower latency as preferable. This stage also involves only lightweight numerical computation and completes in under one second on a CPU.

\subsection{Comparison with Exhaustive Workflow Evaluation}
\label{appendix:cost_exhaustive_comparison}

A naive alternative to FlowCompile is to evaluate every workflow configuration end-to-end and then extract the empirical accuracy--latency frontier. This is computationally infeasible in our setting. The full design spaces contain approximately $2.37$B, $4.84$B, $2.56$M, and $1.04$M workflow configurations for GSM8K, MATH-500, LiveCodeBench, and HotpotQA, respectively. FlowCompile avoids exhaustive workflow-level evaluation by replacing it with one-time sub-agent profiling and analytic workflow-level composition.

The difference is substantial in practice. FlowCompile profiles only $325$, $375$, $225$, and $240$ sub-agent settings in total for GSM8K, MATH-500, LiveCodeBench, and HotpotQA, respectively, and this profiling stage takes about one hour per benchmark. In contrast, on HotpotQA, one full-workflow evaluation takes approximately $10$ minutes. Exhaustively evaluating the full HotpotQA configuration space would therefore require roughly $1.04\text{M} \times 10$ minutes, or about $\textbf{173{,}333}$ \textbf{hours}, whereas FlowCompile completes the benchmark in about \textbf{one hour}.

FlowCompile further reduces the analytic search space through exact sub-agent-level Pareto filtering before workflow-level composition. This pruning is not heuristic: under the monotone accuracy and latency composition rules used by the proxy, any workflow configuration that uses a locally dominated sub-agent configuration is itself dominated by replacing that sub-agent choice with a Pareto-superior one. In practice, this reduces each sub-agent role from roughly $65$--$80$ settings to $6$--$19$ settings. After pruning, the final workflow-level search spaces contain $146{,}751$, $1{,}074{,}866$, $13{,}319$, and $4{,}017$ configurations for GSM8K, MATH-500, LiveCodeBench, and HotpotQA, respectively. Since the remaining search is purely analytic, it completes in $0.16$s, $0.99$s, $0.05$s, and $0.01$s on a CPU for the four benchmarks.

\subsection{Scalability Discussion}
\label{appendix:cost_scalability}

The main scalability advantage of FlowCompile is that expensive model inference is performed at the sub-agent level rather than at the workflow-configuration level. For a fixed profiling set, the profiling cost scales with the number of sub-agent roles and candidate model--budget settings, i.e., $A M R$ in the homogeneous case or $\sum_{a\in\mathcal{A}}|\mathcal{Q}_a|$ in the heterogeneous case. In contrast, the full workflow design space grows combinatorially. Even without structural choices, independently assigning one of $MR$ model--budget settings to each of $A$ sub-agent roles yields $(MR)^A$ workflow configurations; optional structural choices further enlarge this space. FlowCompile avoids this combinatorial inference cost by profiling each sub-agent choice once and reusing the cached profiles to estimate many workflow-level configurations through lightweight numerical composition.

This separation makes the method scalable in practice. The expensive profiling calls are independent across sub-agents, examples, models, and reasoning budgets, and can therefore be parallelized across accelerators. Once collected, the profiles are reused across all workflow-level configurations, deployment preferences, and latency budgets. The remaining workflow-level composition and trade-off set construction require only lightweight numerical computation over cached profiles and complete in under one second on a CPU in our experiments.

Overall, FlowCompile shifts workflow optimization from combinatorial end-to-end evaluation to linear, reusable, and parallelizable sub-agent profiling. This makes compile-time exploration scalable even when the full workflow design space contains millions or billions of configurations.

\begin{figure}[t]
    \centering
    \includegraphics[width=0.6\linewidth]{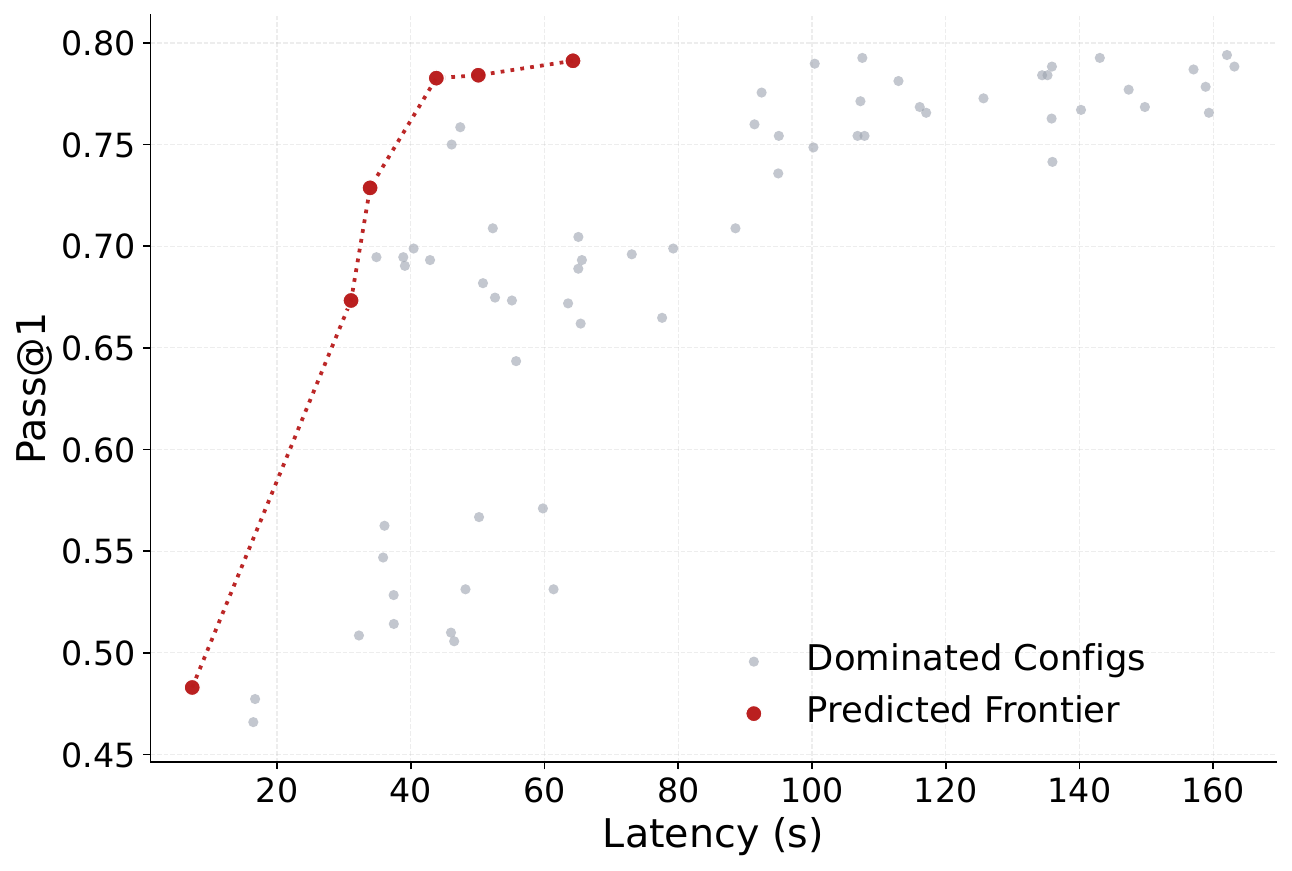}
    \caption{\textbf{Frontier consistency on LiveCodeBench under exhaustive evaluation.}
    The proxy-estimated frontier recovers high-quality empirically measured configurations across the accuracy--latency trade-off space, while most remaining configurations are dominated after full workflow execution.}
    \label{fig:livecodebench_bruteforce}
\end{figure}

\section{Additional Details on Frontier Consistency under Exhaustive Evaluation}
\label{appendix:frontier_consistency_bruteforce}

To further validate the frontier-consistency assumption in Section~\ref{sec:proxy-validation}, we conduct exhaustive-evaluation experiments on HotpotQA and LiveCodeBench under restricted configuration spaces where all workflow configurations can be evaluated end-to-end.
We then compare the proxy-estimated frontier with the empirically measured accuracy--latency trade-off space.

\noindent\textbf{Restricted design space.}
For HotpotQA, we restrict the model choices to Qwen3-1.7B, Qwen3-8B, and Qwen3-14B, and the reasoning budgets to 10 and 10{,}000 tokens.
We consider two workflow structures: the full workflow and a reduced workflow with one generator and one formatter.
This yields $252$ workflow configurations in total.
For LiveCodeBench, we restrict the model choices to Qwen3-4B and Qwen3-8B, and the reasoning budgets to 10 and 10{,}000 tokens.
We consider two workflow structures: the full workflow and a reduced workflow with a single programmer.
This yields $80$ workflow configurations in total.
These restricted spaces preserve the key design dimensions of the full search space, including model choice, reasoning budget, and workflow structure, while making exhaustive end-to-end evaluation feasible.

\noindent\textbf{Results.}
As shown in Figure~\ref{fig:bruteforce_exp}, the proxy-estimated frontier on HotpotQA matches the empirically measured frontier under exhaustive evaluation.
Figure~\ref{fig:livecodebench_bruteforce} shows a similar pattern on LiveCodeBench: the configurations selected by the proxy form a high-quality empirical frontier after workflow execution.
Most configurations outside the proxy frontier are dominated in the measured accuracy--latency space, while the selected configurations cover the main trade-off regimes from low-latency to high-accuracy operating points.
Together, these results provide additional evidence that the workflow-level proxy preserves the non-dominated region well enough to guide compile-time search.

\begin{table}[t]
\centering
\small
\caption{\textbf{Representative accuracy--latency operating points.}
FlowCompile outputs a set of optimized configurations, from which we report accuracy-priority and latency-priority selections. The former targets accuracy comparable to the full Qwen3-14B workflow baseline with lower latency, while the latter targets larger latency reductions while retaining strong accuracy.}
\label{tab:accuracy_latency_tradeoff}
\resizebox{\linewidth}{!}{
\begin{tabular}{
l
S[table-format=2.2] S[table-format=3.1]
S[table-format=2.2] S[table-format=3.1]
S[table-format=2.2] S[table-format=2.1]
S[table-format=2.2] S[table-format=3.1]
}
\toprule
\multirow{2}{*}{\textbf{Method}}
& \multicolumn{2}{c}{\textbf{GSM8K}}
& \multicolumn{2}{c}{\textbf{MATH-500}}
& \multicolumn{2}{c}{\textbf{HotpotQA}}
& \multicolumn{2}{c}{\textbf{LiveCodeBench}} \\

& {Accuracy} & {Latency (s)}
& {Accuracy} & {Latency (s)}
& {F1} & {Latency (s)}
& {Pass@1} & {Latency (s)} \\
\midrule
Full Baseline (Qwen3-14B)
& 95.55 & 168.3
& 96.52 & 355.5
& 86.29 & 26.5
& 80.97 & 329.3 \\

Qwen3-4B
& 95.64 & 94.4
& 93.78 & 187.4
& 83.89 & 12.5
& 78.69 & 186.2 \\

Qwen3-8B
& 95.17 & 134.8
& \best{96.77} & 352.3
& 86.58 & 18.5
& \best{79.69} & 267.2 \\

MaAS (Qwen3-4B)
& 92.42 & 30.5
& 71.82 & 55.0
& 71.82 & 6.3
& 26.56 & 33.7 \\

MaAS (Qwen3-8B)
& 92.80 & 47.9
& 70.40 & 87.1
& 83.44 & 14.1
& 15.20 & \best{16.3} \\

KNN Router
& 89.00 & 42.5
& 78.61 & 82.0
& 73.75 & 6.9
& 50.14 & 141.3 \\

\midrule
\textbf{Ours (Acc.-first)}
& \best{96.02} & 88.3
& 95.27 & 185.5
& \best{86.69} & 7.8
& 78.98 & 51.5 \\

\textbf{Ours (Lat.-first)}
& 92.13 & \best{10.9}
& 85.32 & \best{31.9}
& 85.32 & \best{3.8}
& 41.34 & 19.0 \\
\bottomrule
\end{tabular}
}
\end{table}

\section{Additional Details on Accuracy--Latency Trade-offs}
\label{appendix:accuracy_latency_tradeoff}

Table~\ref{tab:accuracy_latency_tradeoff} provides numerical accuracy and latency results, enabling a direct comparison of FlowCompile and the baselines in the accuracy--latency trade-off space. Since FlowCompile outputs a compiled set of configurations rather than a single operating point, we report two representative selections from this set. The accuracy-priority selection aims to match the full Qwen3-14B workflow baseline while reducing latency, whereas the latency-priority selection minimizes latency while preserving strong task performance.

Across benchmarks, the accuracy-priority configuration achieves accuracy comparable to the full baseline with substantially lower latency.
For example, on HotpotQA, it improves F1 from $86.29$ to $86.69$ while reducing latency from $26.5$s to $7.8$s.
On LiveCodeBench, it reduces latency from $329.3$s to $51.5$s while maintaining similar Pass@1.
The latency-priority configuration yields larger speedups, especially on GSM8K, MATH-500, and HotpotQA, while retaining competitive accuracy.
These results complement the frontier plots in Figure~\ref{fig:pareto_validity} by showing concrete operating points selected from the compiled configuration set.

\begin{figure}[t]
    \centering
    \begin{subfigure}[t]{0.45\linewidth}
        \centering
        \includegraphics[width=\linewidth]{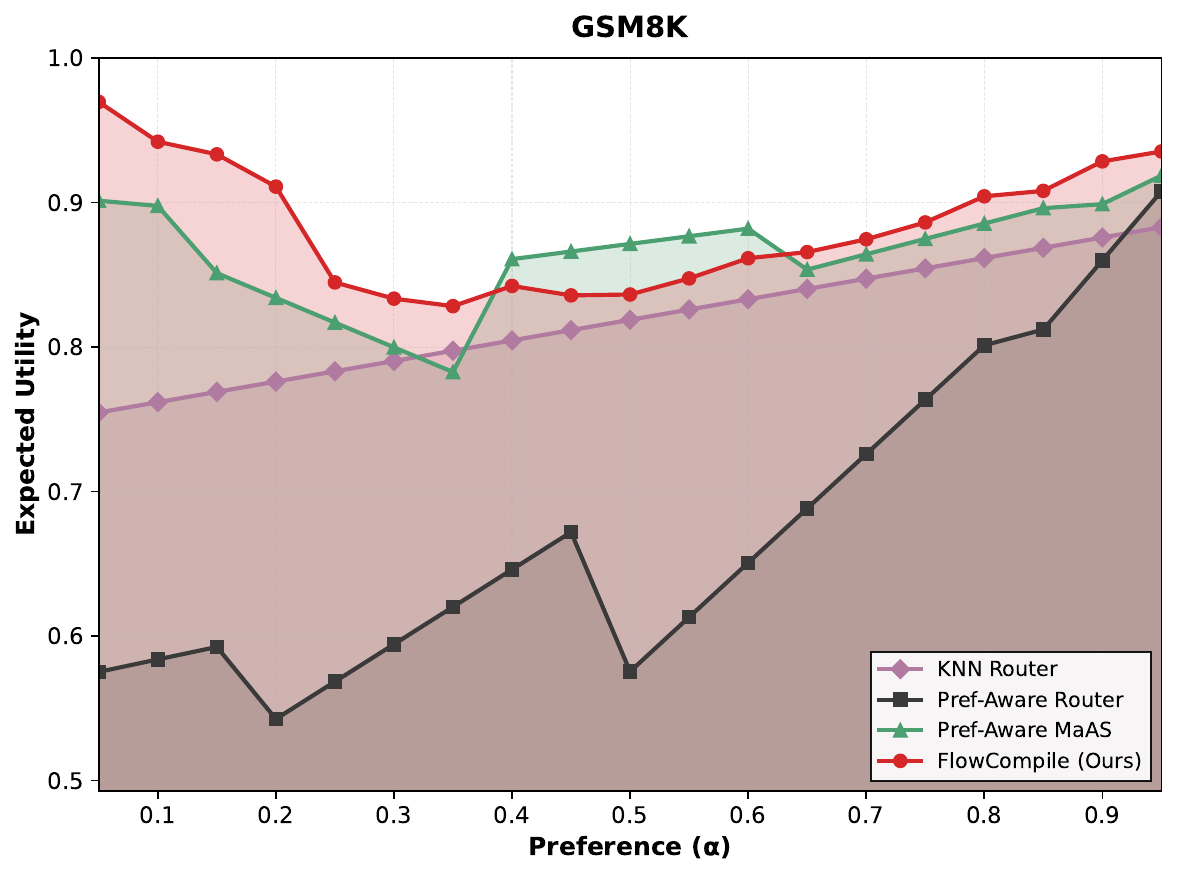}
    \end{subfigure}
    \hfill
    \begin{subfigure}[t]{0.45\linewidth}
        \centering
        \includegraphics[width=\linewidth]{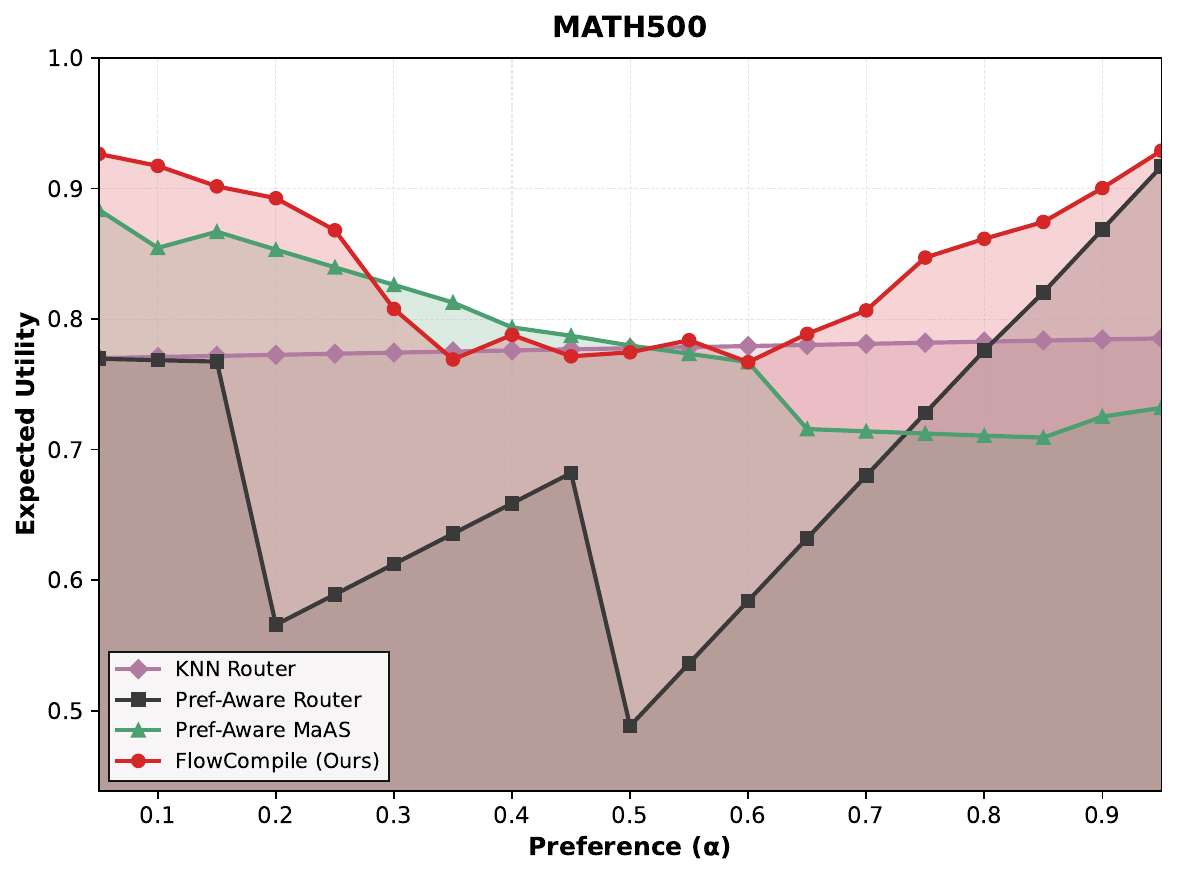}
    \end{subfigure}

    \vspace{0.5em}

    \begin{subfigure}[t]{0.45\linewidth}
        \centering
        \includegraphics[width=\linewidth]{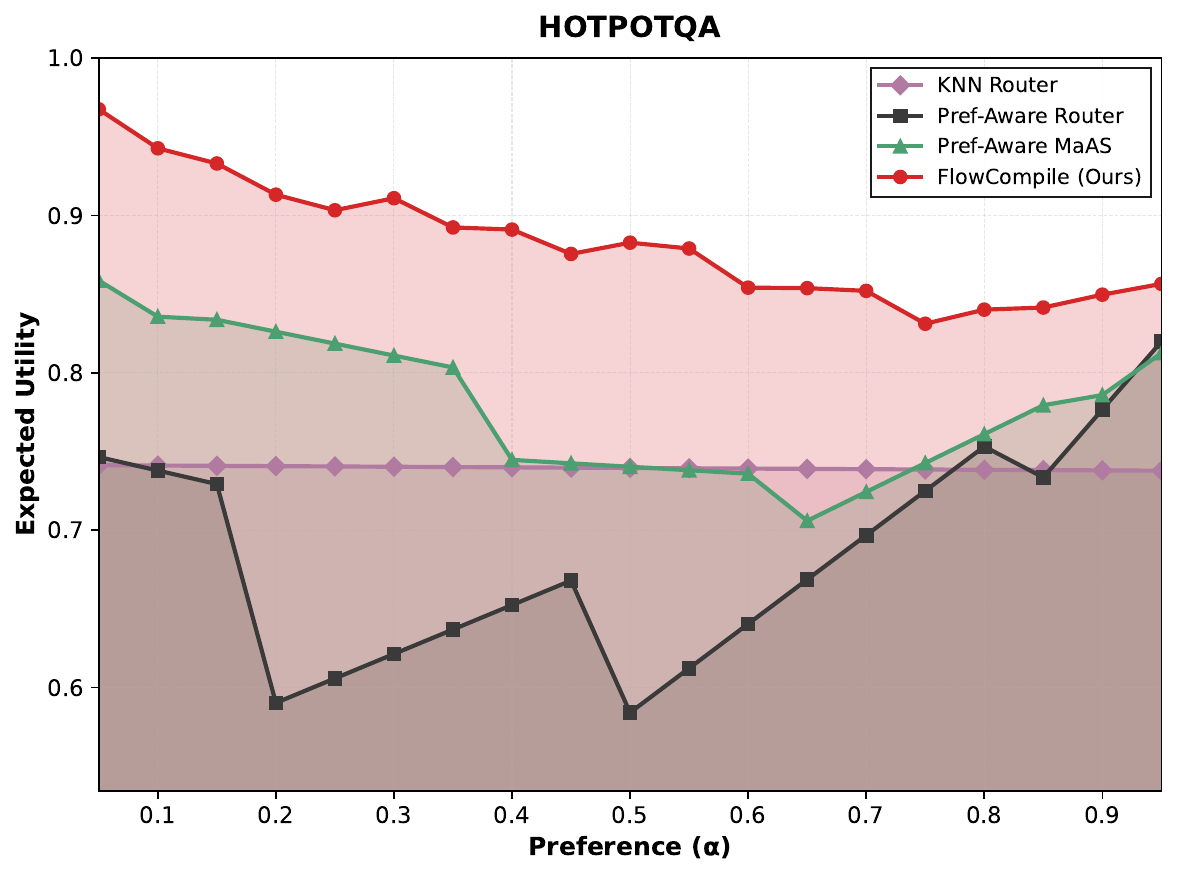}
    \end{subfigure}
    \hfill
    \begin{subfigure}[t]{0.45\linewidth}
        \centering
        \includegraphics[width=\linewidth]{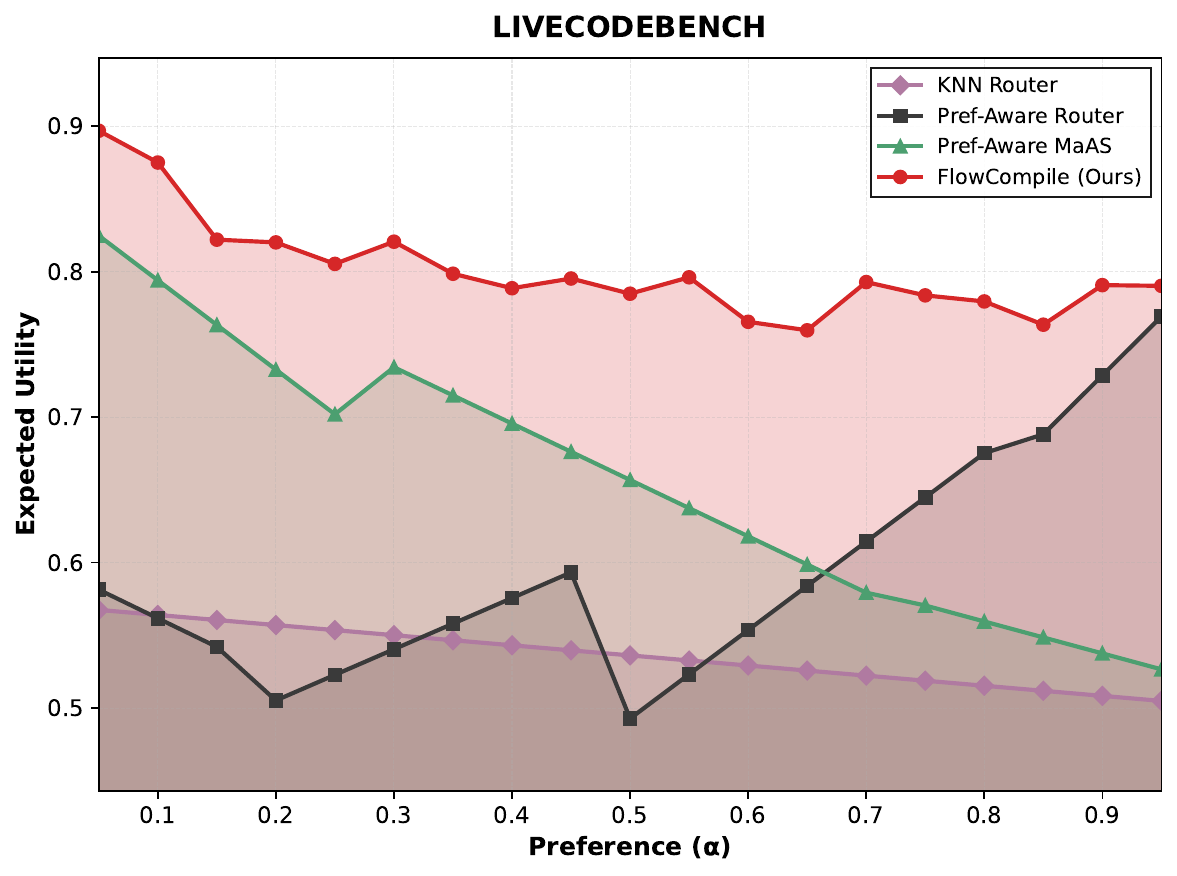}
    \end{subfigure}

    \caption{Per-benchmark average expected utility under fixed accuracy–latency preferences.}
    \label{fig:per_benchmark_fixed_pref_results}
\end{figure}

\section{Additional Details on Preference-Aware Evaluation}
\label{appendix:per_benchmark_eu}

We provide additional details on the preference-aware evaluation protocol.
For FlowCompile, configurations are selected from the compiled set using proxy-estimated accuracy and latency, while reported utilities are computed only from measured test-set performance, avoiding test-set leakage.

\noindent\textbf{Heterogeneous preferences.}
In the heterogeneous setting used in Table~\ref{tab:main_results}, each test query is assigned an independently sampled preference parameter $\alpha \sim \mathrm{Uniform}(0,1)$.
This setting represents mixed workloads containing both latency-sensitive and accuracy-sensitive requests.
For each query, FlowCompile selects the compiled configuration that maximizes proxy-estimated utility under the sampled $\alpha$, and the reported utility is computed using the selected configuration's measured test-set accuracy and latency.
We repeat the sampling ten times and report the mean and standard deviation.

\noindent\textbf{Fixed preferences.}
In the fixed-preference setting, all queries share the same $\alpha$.
We sweep $\alpha$ from $0.05$ to $0.95$ with a step size of $0.05$ to evaluate how robust each method is under different global accuracy--latency preferences.
For each $\alpha$, FlowCompile selects the compiled configuration that maximizes proxy-estimated utility and reports its measured test-set utility.
Figure~\ref{fig:per_benchmark_fixed_pref_results} reports the per-benchmark results for this setting.

Across benchmarks and most preference values, FlowCompile achieves the highest expected utility, indicating that the compiled trade-off set provides robust operating points across diverse accuracy--latency preferences.
Although Pref-Aware MaAS is competitive at a few isolated $\alpha$ values on GSM8K and MATH-500, FlowCompile achieves the best overall performance across the preference spectrum.

\section{Additional Details and Analysis on Per-Query Routing}
\label{appendix:per_query_results}

\noindent\textbf{Setup and main result.}
FlowCompile performs compile-time optimization and produces a global accuracy--latency trade-off set. This compiled set is complementary to runtime routing: instead of routing over the full combinatorial design space, a router only needs to select among the compact set of configurations produced by FlowCompile.

We evaluate this use case on GSM8K and MATH-500 using a simple KNN router with $k=20$. For each test query, the router uses Longformer~\citep{beltagy2020longformer} embeddings to retrieve nearest neighbors from the validation set, and then selects a configuration from the compiled set based on the neighbors' performance. This adds only lightweight query-level selection and requires no additional workflow profiling, online optimization, or retraining.

As shown in Table~\ref{tab:per-query-routing}, routing over the compiled set further improves expected utility from $89.2$ to $91.8$ on GSM8K and from $84.2$ to $90.5$ on MATH-500. These results show that FlowCompile can be used either directly as a compiled deployment artifact or as a high-quality candidate pool for downstream per-query adaptation.

\begin{table}[t]
\centering
\small
\caption{\textbf{Per-query routing over compiled configurations.}
A lightweight KNN router further improves expected utility by selecting among the configurations produced by FlowCompile for each query.}
\label{tab:per-query-routing}
\begin{tabular}{lcc}
\toprule
\textbf{Method} & \textbf{GSM8K} & \textbf{MATH-500} \\
\midrule
FlowCompile & $89.2 \pm 0.6$ & $84.2 \pm 0.8$ \\
FlowCompile + KNN Routing & $\mathbf{91.8 \pm 0.5}$ & $\mathbf{90.5 \pm 0.9}$ \\
\bottomrule
\end{tabular}
\end{table}

\noindent\textbf{Analysis setup.}
To better understand how query-level selection uses the compiled set, we further analyze the routing assignments on MATH-500, since it provides difficulty labels that support post hoc interpretation. The difficulty labels are not used by the router during selection; they are used only to examine whether the selected configurations correlate with problem difficulty. For each test query and preference value $\alpha$, the assignment artifact records the selected compiled workflow configuration, including its workflow structure, sub-agent model choices, and reasoning budgets.

We group preference values into three regimes: Latency-first ($\alpha \leq 0.3$), Balanced ($0.4 \leq \alpha \leq 0.6$), and Accuracy-first ($\alpha \geq 0.7$). We also classify the selected workflow structures into Simple, Balanced, and Complex classes, corresponding to two, three, and four active sub-agents in the MATH-500 workflow, respectively.

\begin{figure*}[t]
\centering
\includegraphics[width=\textwidth]{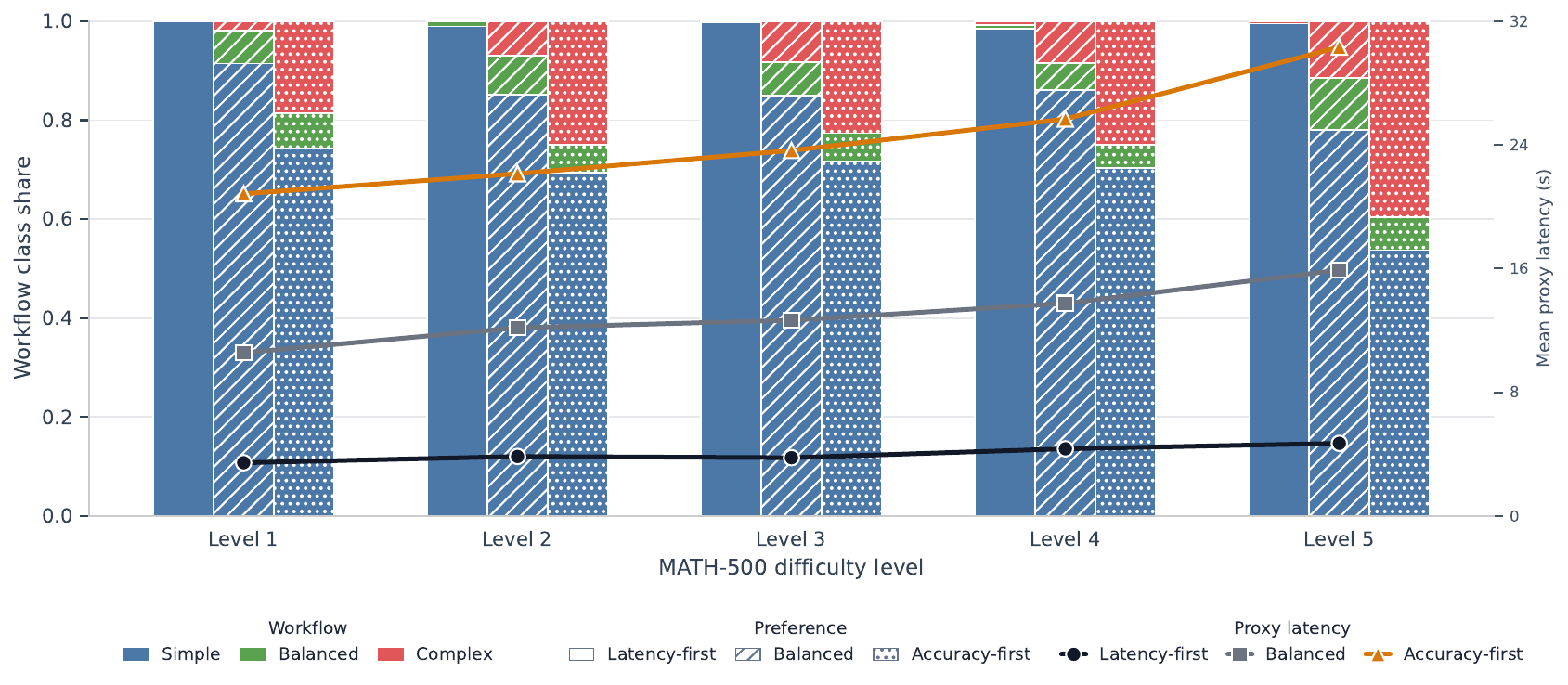}
\caption{\textbf{Routing assignment patterns on MATH-500.}
Difficulty labels are used only for post hoc analysis and are not observed by the router during selection. For each difficulty level, stacked bars show the workflow-class distribution of selected configurations under Latency-first, Balanced, and Accuracy-first preference regimes; hatching distinguishes preference regimes and colors indicate workflow classes. Lines with markers report the corresponding mean proxy latency. The selected configurations remain mostly Simple under Latency-first preferences, but shift more often toward Balanced or Complex workflows for harder questions as accuracy is prioritized.}
\label{fig:math500-routing-config-distribution}
\end{figure*}

\noindent\textbf{Routing assignment patterns.}
Figure~\ref{fig:math500-routing-config-distribution} shows that per-query routing does not select a single global configuration for all MATH-500 problems. Instead, it induces different configuration distributions across difficulty levels and preference regimes, even though difficulty labels are not available to the router. In the Latency-first regime, all difficulty levels are dominated by Simple workflows, and the mean proxy latency remains low. Thus, when latency is prioritized, the router preserves inexpensive execution even for harder questions.

The difficulty-conditioned pattern becomes more pronounced in the Balanced and Accuracy-first preference regimes. In the Balanced regime, the share of Complex workflows increases from $1.9\%$ for level-1 questions to $11.4\%$ for level-5 questions, while mean proxy latency rises from $10.6$s to $15.9$s. In the Accuracy-first regime, the Complex workflow share further increases from $18.6\%$ for level-1 questions to $39.6\%$ for level-5 questions, with mean proxy latency increasing from $20.8$s to $30.3$s. This indicates that the router spends more workflow-level compute on harder questions, but mainly when the preference value makes the additional latency worthwhile.

\begin{figure}[t]
\centering
\includegraphics[width=0.5\linewidth]{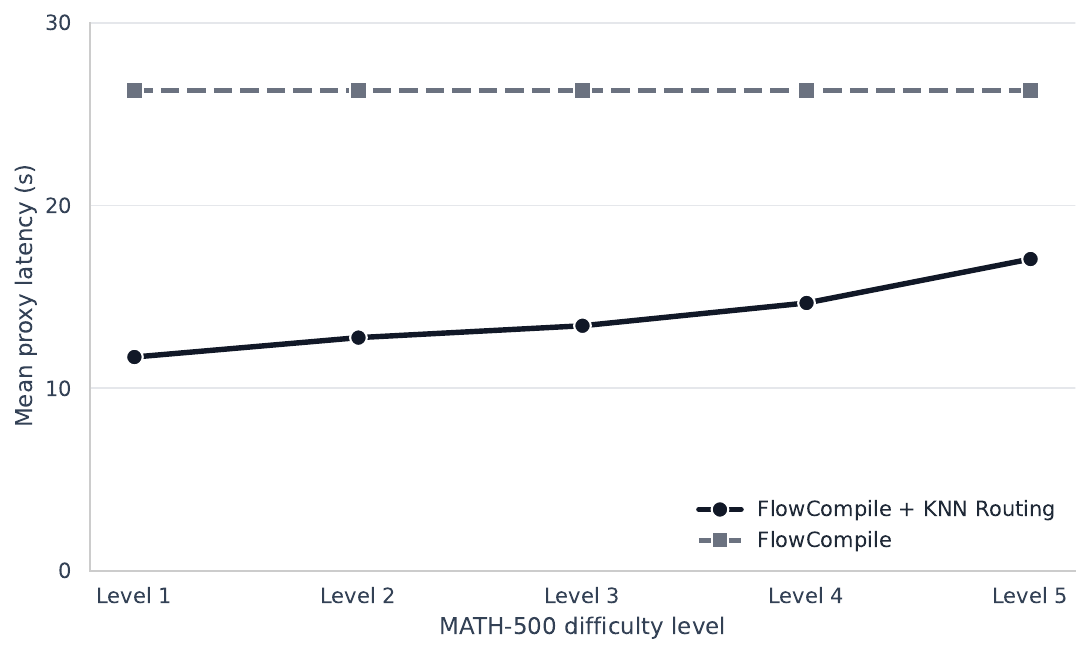}
\caption{\textbf{Mean proxy latency by MATH-500 difficulty level, with and without per-query routing.}
Difficulty labels are used only for post hoc analysis and are not observed by the router during selection. FlowCompile without routing applies a single global configuration, resulting in an approximately constant latency across difficulty levels. In contrast, FlowCompile + KNN Routing produces a difficulty-sensitive latency profile: easier questions are assigned lower-latency configurations, while harder questions receive progressively higher-latency configurations.}
\label{fig:math500-routing-latency-by-difficulty}
\end{figure}

\noindent\textbf{Latency allocation by difficulty.}
Figure~\ref{fig:math500-routing-latency-by-difficulty} provides a complementary view by directly comparing the mean proxy latency across MATH-500 difficulty levels, with and without per-query routing. FlowCompile without routing applies a single global configuration, and therefore operates at essentially the same latency level for all difficulty levels. In contrast, FlowCompile + KNN Routing produces a difficulty-sensitive latency profile: easier questions are assigned cheaper configurations, while harder questions receive progressively more expensive ones.

This comparison shows that per-query routing changes how configurations are selected within the compiled set. Without routing, FlowCompile selects configurations according to the global preference objective, so the selected latency is nearly constant across MATH-500 difficulty levels after grouping by difficulty. In contrast, KNN routing makes query-conditioned selections from the same compiled set: easier questions receive lower-latency configurations, while harder questions receive relatively higher-latency configurations. Although the routed configurations remain lower-latency than the non-routed FlowCompile selection across all difficulty levels, the increasing trend shows that routing decisions correlate with problem difficulty, although difficulty labels are never observed by the router.

\noindent\textbf{Implication.}
Together, these analyses illustrate the complementarity between compile-time optimization and runtime routing. FlowCompile first constructs a compact set of high-quality configurations spanning workflow structures, model choices, and reasoning budgets. The KNN router then performs lightweight query-conditioned selection within this set. Although the router does not observe difficulty labels, its selected configurations correlate with problem difficulty: easier questions are more often assigned to simpler, lower-latency compiled workflows, while harder MATH-500 questions are routed more frequently to configurations with additional reasoning branches and higher latency when extra compute is likely to help. The improvement from per-query routing therefore comes from using the compiled set as a query-conditioned deployment menu, rather than retraining the router or searching the full configuration space online.

\begin{figure}[t]
    \centering
    \includegraphics[width=0.5\linewidth]{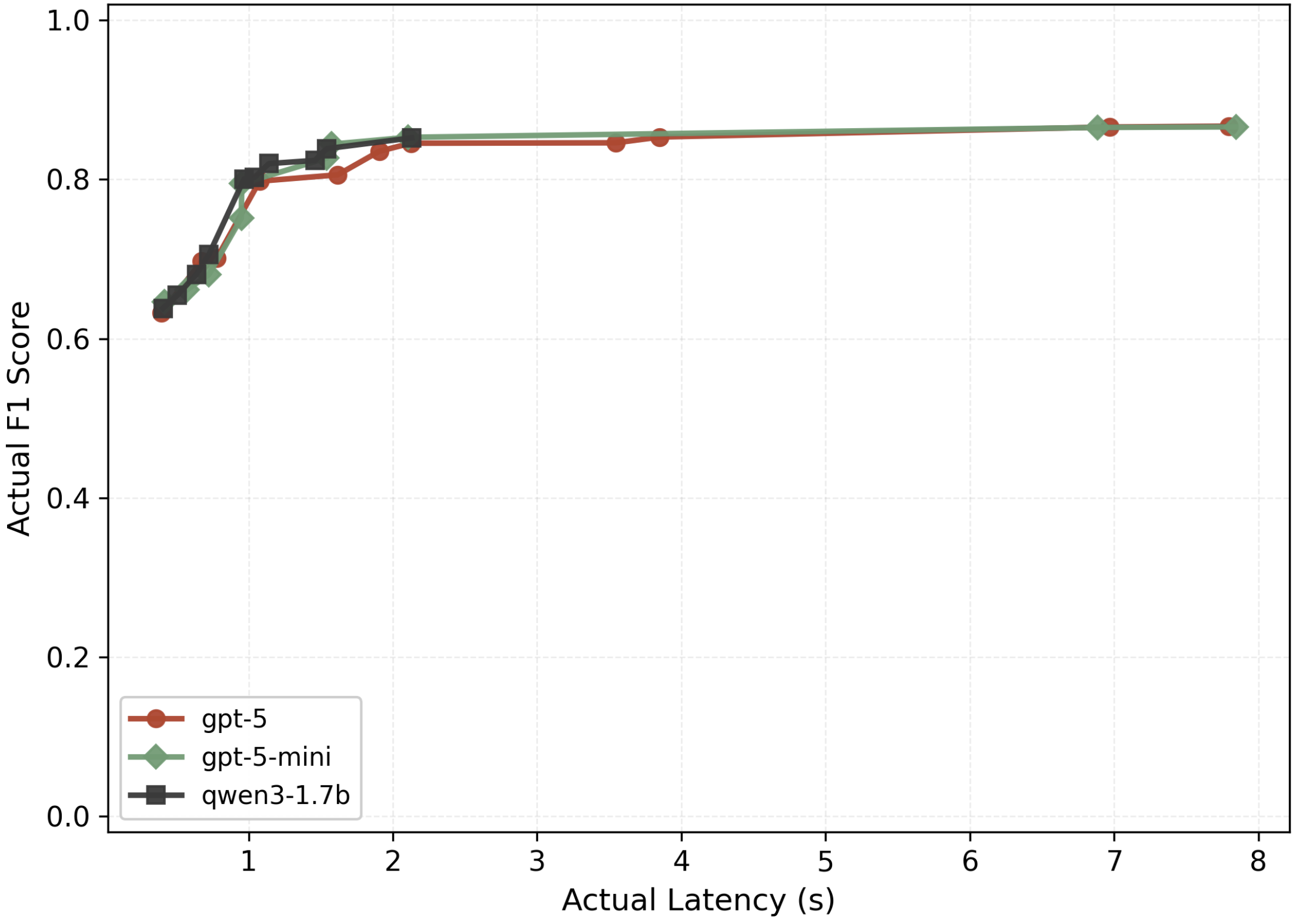}
    \caption{\textbf{Reference-model ablation on HotpotQA.}
    FlowCompile produces similar optimized trade-off sets when sub-agent data are induced using different reference models, suggesting that the compilation pipeline is robust to the reference model choice.}
    \label{fig:reference_model_ablation}
\end{figure}

\section{Additional Details on Reference-Model Ablation}
\label{appendix:reference_model_ablation}

FlowCompile uses a reference model to induce sub-agent-level profiling data from workflow traces.
We evaluate whether the compiled trade-off set is sensitive to this reference model choice.
In addition to the default GPT-5 reference model, we repeat the sub-agent data induction, profiling, compositional estimation, and frontier search pipeline on HotpotQA using GPT-5-mini and Qwen3-1.7B as alternative reference models.

Figure~\ref{fig:reference_model_ablation} shows that the resulting optimized configuration sets achieve largely consistent measured performance across reference models.
This suggests that FlowCompile is not tightly coupled to a particular high-capacity reference model, provided that the reference model can generate reasonable workflow traces for sub-agent profiling.

\begin{table*}[t]
\centering
\small
\caption{\textbf{Qualitative structure of compiled workflow configurations.}
We summarize dominant low-latency and high-accuracy patterns using Simple, Balanced, and Complex workflow classes.}
\label{tab:compiled-config-patterns}
\setlength{\tabcolsep}{4pt}
\renewcommand{\arraystretch}{1.15}
\begin{tabular}{
    p{0.14\textwidth}
    p{0.24\textwidth}
    p{0.26\textwidth}
    p{0.27\textwidth}
}
\toprule
\textbf{Benchmark}
& \textbf{Latency-priority pattern}
& \textbf{Accuracy-priority pattern}
& \textbf{Main trade-off mechanism} \\
\midrule
GSM8K
& Mostly Simple workflows with short budgets and smaller models.
& Mostly Complex workflows with larger budgets and stronger solving or aggregation models.
& FlowCompile activates additional math reasoning stages and allocates more capacity to solving and aggregation. \\

\addlinespace[2pt]
MATH-500
& Mostly Simple workflows with modest budgets.
& Mostly Complex workflows with substantially larger budgets.
& FlowCompile spends latency on additional reasoning paths for harder math problems and shifts capacity toward stronger solving and aggregation calls. \\

\addlinespace[2pt]
HotpotQA
& A Simple fixed-size workflow with smaller models and short budgets.
& The same Simple workflow with stronger model assignments and larger budgets.
& The workflow structure remains stable; the trade-off is mainly governed by model and budget allocation within answer generation and formatting. \\

\addlinespace[2pt]
LiveCodeBench
& Repair-based workflows with short generation and repair budgets in the low-latency region.
& Complex repair workflows with substantially larger code-generation and repair budgets.
& Accuracy is improved primarily by allocating more compute and stronger models to code generation and execution-guided repair. \\
\bottomrule
\end{tabular}
\end{table*}

\begin{figure*}[t]
\centering
\includegraphics[width=0.8\textwidth]{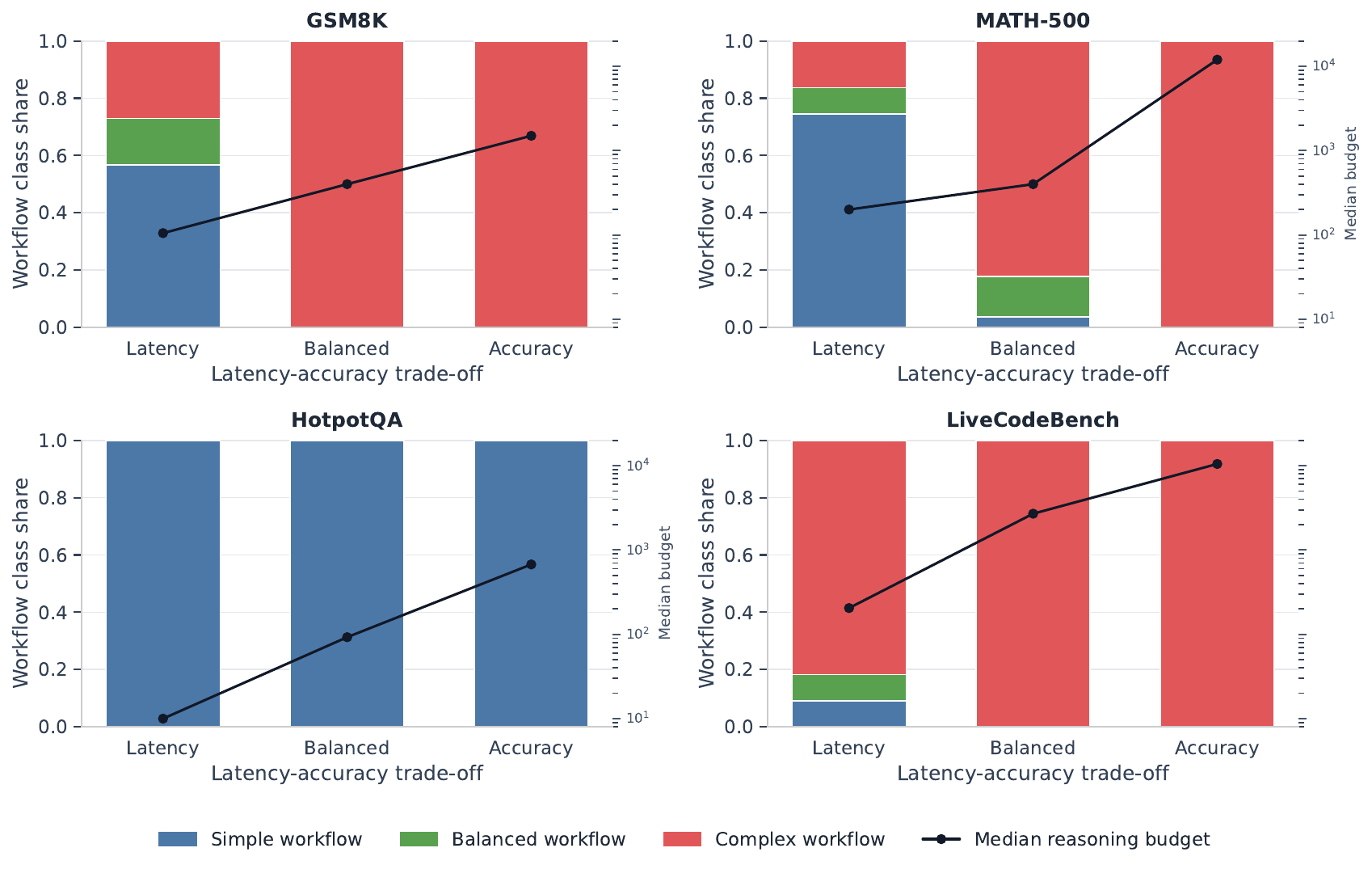}
\caption{Workflow-structure patterns across latency-ranked regions of the full compiled artifacts. For each benchmark, configurations are grouped into low-latency, middle, and high-latency regions. Stacked bars show the proportion of Simple, Balanced, and Complex workflow classes, and black markers show the median reasoning budget across configured sub-agent calls on a log scale.}
\label{fig:compiled-config-terciles}
\end{figure*}

\begin{figure*}[t]
\centering
\includegraphics[width=0.8\textwidth]{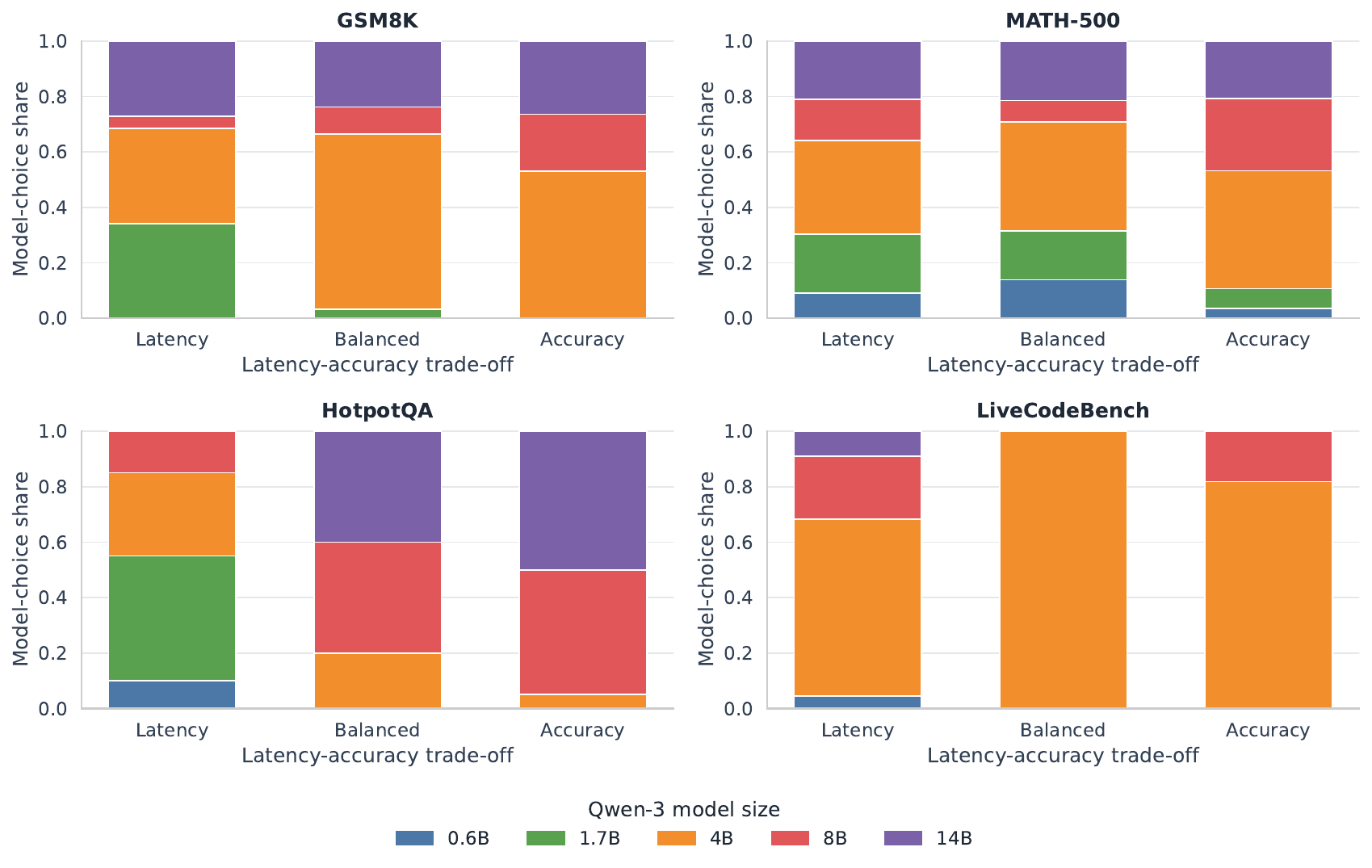}
\caption{Model choices across latency-ranked regions of the full compiled Pareto artifacts. Stacked bars show the distribution of Qwen-3 model sizes used by configured sub-agent calls in each region. Moving from latency-priority to accuracy-priority configurations generally shifts capacity from smaller and mid-sized models toward 8B and 14B models, with task-dependent allocation patterns.}
\label{fig:compiled-config-models}
\end{figure*}

\section{Analysis of Compiled Workflow Configurations}
\label{appendix:analysis-compiled-workflow-configs}

We further inspect the compiled configuration sets for all four benchmarks to characterize how FlowCompile allocates workflow structure, reasoning budget, and model capacity across different accuracy--latency operating regimes.

\noindent\textbf{Analysis setup.}
We group compiled configurations into three accuracy--latency trade-off regions: Latency-priority, Balanced-priority, and Accuracy-priority, corresponding to low-latency, intermediate, and high-accuracy regions of the compiled set, respectively. These trade-off regions describe where a configuration lies on the compiled accuracy--latency frontier and are distinct from workflow-structure classes.

We also use a benchmark-specific three-way structural classification. For GSM8K and MATH-500, Simple, Balanced, and Complex correspond to configurations with two, three, and four active sub-agents, respectively. For LiveCodeBench, the number of active sub-agents in the compiled set remains fixed across configurations, so Simple, Balanced, and Complex correspond to one, two, and three maximum repair attempts. For HotpotQA, the compiled set uses a fixed two-sub-agent workflow, so the structural class remains Simple, and the main variation comes from model and budget allocation.

\noindent\textbf{Overall pattern.}
Table~\ref{tab:compiled-config-patterns} summarizes the dominant patterns for each benchmark, while Figures~\ref{fig:compiled-config-terciles} and~\ref{fig:compiled-config-models} provide aggregated views of workflow structure, reasoning budget, and model choices across the three trade-off regions. Across benchmarks, the compiled configurations exhibit interpretable structure rather than arbitrary combinations of models, budgets, and workflow variants. Moving from Latency-priority to Accuracy-priority configurations is mainly explained by three changes: the workflow becomes more complex when additional reasoning stages are useful, reasoning budgets increase for task-critical sub-agents, and model assignments shift toward stronger models. Low-latency configurations therefore tend to use simpler workflows, smaller or mid-sized models, and short reasoning budgets, while high-accuracy configurations allocate more compute to bottleneck roles such as solving, aggregation, formatting, refinement, or repair.

\noindent\textbf{Math reasoning workflows.}
GSM8K and MATH-500 show the clearest structural transition. In the latency-priority region, FlowCompile often selects Simple configurations. As the objective shifts toward accuracy, the compiled set increasingly favors Complex configurations. This suggests that, for math reasoning, accuracy is not obtained only by uniformly scaling every LLM call. Instead, FlowCompile selectively activates additional reasoning stages and allocates larger budgets or stronger models to stages that provide complementary reasoning signals.

\noindent\textbf{Multi-hop QA workflows.}
HotpotQA follows a different pattern. The compiled configurations keep the same Simple workflow class across the trade-off range. Low-latency configurations use smaller models and shorter budgets, while high-accuracy configurations increase capacity within the same workflow structure. This suggests that, for this workflow, the dominant optimization choice is not whether to activate additional sub-agents, but how much capacity to allocate to answer generation and final formatting.

\noindent\textbf{Code reasoning workflows.}
LiveCodeBench also keeps a stable set of active sub-agents, but exposes a different structural knob through the number of execution-guided repair attempts. Latency-priority configurations use shorter generation and repair budgets, while balanced and accuracy-priority configurations are dominated by more repair trials. The main difference between balanced and accuracy-priority regions is therefore budget and model allocation: when accuracy is prioritized, FlowCompile spends substantially more compute on producing and repairing executable code.

\paragraph{Implication.}
Together, Table~\ref{tab:compiled-config-patterns}, Figure~\ref{fig:compiled-config-terciles} and Figure~\ref{fig:compiled-config-models} show that the compiled set can be interpreted as a deployment menu rather than a collection of unrelated operating points. Latency-priority configurations remove optional reasoning stages or use small-budget versions of the same task skeleton, while accuracy-priority configurations spend compute where the workflow benefits most: additional reasoning paths for math, stronger generation and formatting for question answering, and more repair attempts together with larger generation and repair budgets for code reasoning. This supports the compiler view of FlowCompile: its output is a reusable set of workflow-level configurations spanning qualitatively different operating regimes.

\section{Metric Description}
\label{appendix:metric_desc}

\subsection{Expected Utility}
We use expected utility to summarize the accuracy--latency trade-off under different deployment preferences. For a workflow configuration $c$, let $\mathrm{Acc}(c)$ denote its task performance and $\mathrm{Lat}(c)$ denote its measured latency. We first convert latency into a latency-efficiency score,
\begin{equation}
\mathrm{LES}(c)
=
1 - \frac{\mathrm{Lat}(c)}{L_{\max}},
\end{equation}
where $L_{\max}$ is the maximum latency among the configurations and baselines being compared under the same benchmark. Expected utility is then defined as
\begin{equation}
U(c;\alpha)
=
\alpha \, \mathrm{Acc}(c)
+
(1-\alpha)\,\mathrm{LES}(c),
\end{equation}
where $\alpha \in (0,1)$ controls the accuracy--latency preference. Smaller $\alpha$ places more weight on latency efficiency, while larger $\alpha$ places more weight on accuracy. For QA tasks, $\mathrm{Acc}(c)$ denotes the F1 score; for LiveCodeBench, it denotes Pass@1.

For preference-aware evaluation, configuration selection and test-set evaluation are separated.
For methods that output a fixed single configuration, we directly compute its measured test-set
utility \(U(c;\alpha)\). For methods that provide a candidate set, including FlowCompile and
preference-aware baselines, the configuration is selected using validation/proxy estimates:
\[
\hat{c}_{\alpha}
=
\arg\max_{c \in S}
\widehat{U}(c;\alpha),
\]
where \(S\) denotes the candidate configuration set available to the method and
\(\widehat{U}\) is computed from estimated accuracy and latency. We then report the measured
test-set utility of the selected configuration:
\[
U(S;\alpha)
=
U(\hat{c}_{\alpha};\alpha).
\]
For FlowCompile, \(S=\widehat{\mathcal{F}}\) is the compiled configuration set. No test-set
measurements are used for configuration selection.

In the heterogeneous-preference setting, $\alpha$ is sampled independently for each query and results are averaged across repeated samples. In the fixed-preference setting, a single $\alpha$ is shared across all queries and swept over $(0,1)$.

\subsection{Proxy Validation Metrics}
Let $\mathcal{C} = \{c_1, \ldots, c_N\}$ denote the set of evaluated workflow configurations.
For each configuration $c_i$, let $\hat{y}_i$ and $y_i$ denote the estimated and empirically measured values of a given metric, respectively.

\noindent\textbf{Spearman Correlation.}
Spearman correlation measures the consistency between the relative orderings induced by the estimated and measured values.
Let $\mathrm{rank}(\hat{y}_i)$ and $\mathrm{rank}(y_i)$ denote the ranks of $\hat{y}_i$ and $y_i$ among all configurations in $\mathcal{C}$.
The Spearman rank correlation coefficient is defined as
\begin{equation}
\rho
=
\frac{\sum_{i=1}^{N}
\left(\mathrm{rank}(\hat{y}_i) - \overline{\mathrm{rank}(\hat{y})}\right)
\left(\mathrm{rank}(y_i) - \overline{\mathrm{rank}(y)}\right)}
{\sqrt{
\sum_{i=1}^{N}\left(\mathrm{rank}(\hat{y}_i) - \overline{\mathrm{rank}(\hat{y})}\right)^2
\sum_{i=1}^{N}\left(\mathrm{rank}(y_i) - \overline{\mathrm{rank}(y)}\right)^2
}},
\end{equation}
where $\overline{\mathrm{rank}(\hat{y})}$ and $\overline{\mathrm{rank}(y)}$ denote the mean ranks.
Spearman correlation takes values in $[-1,1]$, with higher values indicating stronger agreement in relative ordering.

\noindent\textbf{Pairwise Agreement.}
Pairwise agreement directly measures the fraction of configuration pairs whose relative ordering is correctly preserved.
Formally, it is defined as
\begin{equation}
\mathrm{PA}
=
\frac{1}{\binom{N}{2}}
\sum_{1 \le i < j \le N}
\mathbb{I}
\left[
\operatorname{sign}(\hat{y}_i - \hat{y}_j)
=
\operatorname{sign}(y_i - y_j)
\right],
\end{equation}
where $\mathbb{I}[\cdot]$ is the indicator function.
Pairwise agreement lies in $[0,1]$ and equals $1$ if and only if the estimated values induce exactly the same total ordering as the measured values.

\noindent\textbf{Calibrated MAE.}
Raw mean absolute error (MAE) is sensitive to systematic discrepancies between estimated and measured values, including both global offsets and scale mismatches. Since our objective is to evaluate how well the compiler captures configuration-dependent accuracy--latency trade-offs, we apply an affine calibration to remove systematic bias and scale distortion before computing MAE. We use a two-point calibration rather than fitting an affine regressor over all measured configurations because the latter would require measuring many workflow configurations, whereas our goal is to assess calibrated absolute error under a minimal-calibration setting consistent with proxy-based compilation.

Formally, let $i_{\min} = \arg\min_i \hat{y}_i$ and $i_{\max} = \arg\max_i \hat{y}_i$ denote the indices of the configurations with the minimum and maximum estimated values, respectively. We define
\begin{equation}
\hat{y}_{\min} = \hat{y}_{i_{\min}}, \quad
\hat{y}_{\max} = \hat{y}_{i_{\max}}, \quad
y_{\min} = y_{i_{\min}}, \quad
y_{\max} = y_{i_{\max}}.
\end{equation}
That is, the calibration anchors are determined by the extrema of the predicted values, and the corresponding measured values are taken from the same configurations.

We then define an affine mapping that aligns these two anchor points:
\begin{equation}
\hat{y}_i^{\,\mathrm{cal}}
=
\frac{y_{\max} - y_{\min}}{\hat{y}_{\max} - \hat{y}_{\min}}
\left(\hat{y}_i - \hat{y}_{\min}\right)
+ y_{\min}.
\end{equation}
The calibrated MAE is computed as
\begin{equation}
\mathrm{cMAE}
=
\frac{1}{N}
\sum_{i=1}^{N}
\left|
\hat{y}_i^{\,\mathrm{cal}} - y_i
\right|.
\end{equation}

This two-point affine calibration removes global scale and bias effects using only the predicted extrema as anchors, while preserving the relative structure of the estimated values. Although it is less statistically robust than fitting a regression over many measured configurations, it better matches the intended low-calibration-cost setting of FlowCompile. Since the main role of the proxy is to preserve trade-off structure for configuration search, we report cMAE together with rank-based metrics such as Spearman correlation and pairwise agreement.

\section{LLM-as-a-Judge Protocols}
\label{appendix:judge_prompt}

FlowCompile uses LLM-as-a-judge evaluation in two distinct stages. During sub-agent data generation, the judge filters intermediate calls from reference-model workflow traces and retains calls that are well-executed and useful for producing a successful final answer. During isolated sub-agent profiling, the judge evaluates outputs from candidate model--budget configurations by comparing them against the induced pseudo target for the corresponding sub-agent, without observing downstream workflow traces. All judge-based filtering and profiling are performed only on the profile set.

\subsection{Prompt for Sub-Agent Data Generation}
\label{appendix:judge_prompt_data_generation}

\begin{Verbatim}[fontsize=\small, breaklines]
You are an expert evaluator judging whether a specific AI agent call helped solve a problem within a multi-agent workflow.

**Original Problem:**
{agent_data['problem']}

**Ground Truth Answer:**
{agent_data['ground_truth']}

**Workflow Final Answer:**
{final_answer}{quality_info}

**Full Workflow Trace (showing all steps leading to the solution):**
{trace_context}

**Current Agent Being Evaluated:**
- Agent Name: {agent_data['agent_name']}
- Step Number: {agent_data['step_number']}
- Agent-Level Input: {agent_data['agent_input']}

**What the LLM Received (Raw Prompt):**
{agent_data['raw_llm_prompt']}

**What the LLM Generated (Raw Output):**
{agent_data['raw_llm_output']}

**Final Agent Output (after any post-processing):**
{actual_agent_output}

**Evaluation Task:**
Determine if THIS SPECIFIC agent call was helpful and well-executed. Consider:

1. **Understanding**: Did the agent correctly understand its task from the input prompt?
2. **Execution Quality**: Is the agent's raw output (what the LLM generated) sound, logical, and well-reasoned?
3. **Correctness**: Does the output align with the ground truth and contribute to a high-quality final answer?
4. **Value**: In the context of the full workflow, did this agent's contribution help?
5. **Intermediate Utility**: If this is an intermediate step, did it provide useful information for subsequent agents?

**Important Notes:**
- The overall workflow produced a HIGH-QUALITY answer
- Your job is to assess if THIS PARTICULAR agent call was done well
- An agent can be marked correct even if it's an intermediate step (e.g., programmer generating code, answer_generate producing one of multiple candidate solutions)
- Focus on: Was the agent's output reasonable, helpful, and properly executed given its role?

Respond in JSON format:
{{
    "is_correct": true/false,
    "reasoning": "Brief explanation (2-3 sentences) of whether this agent call was helpful and well-executed"
}}
\end{Verbatim}

\subsection{Prompt for Isolated Sub-Agent Profiling}
\label{appendix:judge_prompt_profiling}

After sub-agent data generation, FlowCompile profiles each candidate model--budget configuration independently on the induced sub-agent dataset. This stage does not generate or observe downstream workflow traces. For each induced example, the profiled configuration receives the same sub-agent input and produces a candidate output, which is compared against the induced pseudo target output for that sub-agent role. When exact or executable evaluation is available, we use deterministic evaluation instead of an LLM judge, such as normalized answer matching for math, task F1 for HotpotQA final answers, and execution against public test cases for LiveCodeBench programs. The profiling judge is used only when deterministic matching is not applicable, such as for evaluating aggregation outputs.

\begin{Verbatim}[fontsize=\small, breaklines]
You are an expert evaluator for a sub-agent in a structured LLM workflow.

**Original Problem:**
{agent_data['problem']}

**Ground Truth Answer:**
{agent_data['ground_truth']}

**Current Agent Being Evaluated:**
- Agent Name: {agent_data['agent_name']}
- Agent Description: {agent_data['agent_description']}
- Agent-Level Input: {agent_data['agent_input']}

**Pseudo Target Output:**
{pseudo_target_output}

**Candidate Output:**
{candidate_output}

**Evaluation Task:**
Determine whether the Candidate Output is correct for this sub-agent role. A correct output should be semantically equivalent to the Pseudo Target Output or provide the same useful information needed by downstream workflow stages. Consider:

1. **Role Fulfillment**: Does the Candidate Output fulfill the intended role of this sub-agent?
2. **Semantic Correctness**: Is it semantically equivalent to, or at least as useful as, the Pseudo Target Output?
3. **Completeness**: Does it contain the key information needed by downstream stages?
4. **Format Compatibility**: Is the output in a format that can be consumed by the subsequent workflow stage?

**Important Notes:**
- Do not require identical wording.
- Focus on correctness, usefulness, and compatibility with the sub-agent role.
- Do not judge whether the entire workflow would succeed; judge only this sub-agent output relative to its induced pseudo target.
- If the Candidate Output is partially correct but misses key information needed by downstream stages, mark it incorrect.

Respond in JSON format:
{{
    "is_correct": true/false,
    "reasoning": "Brief explanation (1-2 sentences) of whether this candidate output is correct for the sub-agent role"
}}
\end{Verbatim}

The held-out test set is used only for final workflow evaluation and is never used for sub-agent data generation, sub-agent profiling, configuration selection, or judge-based filtering.

\section{Limitations}
\label{appendix:limitations}

FlowCompile is designed for structured LLM workflows whose execution graphs and sub-agent interfaces are specified in advance. This scope matches many program-like workflow systems, but it is less directly applicable to open-ended agentic systems whose execution traces are dynamically constructed at inference time, such as ReAct-style agents with unbounded tool-use patterns~\citep{yao2022react}. Extending workflow compilation to such dynamic settings would require additional mechanisms for trace abstraction or online workflow-graph construction.

FlowCompile also relies on a workflow-level proxy for compile-time search. Although our experiments validate frontier consistency and local order preservation across the evaluated benchmarks, the proxy remains an approximation rather than an exact simulator. Its effectiveness depends on how well independently induced sub-agent profiles capture the dominant interactions among sub-agents, including the intermediate inputs produced by different upstream configurations. Our proxy-validation results suggest that such distribution shifts do not substantially affect the frontier structure in the evaluated workflows, allowing FlowCompile to reliably identify high-quality configurations even if absolute predictions for some low-quality configurations may be less accurate. More explicitly modeling these shifts could further improve the compiler's estimation accuracy and extend its effectiveness to more heterogeneous and complex workflows.

\section{Broader Impact}
\label{appendix:broader_impact}

FlowCompile aims to improve the efficiency of structured LLM workflows by compiling a reusable set of configurations that span accuracy--latency trade-offs. A potential positive impact is that such optimization can reduce inference cost and latency, making structured LLM systems more accessible under practical deployment constraints. By exposing multiple operating points rather than a single configuration, FlowCompile may also help practitioners make more transparent deployment decisions based on resource budgets and task requirements.

At the same time, FlowCompile is an optimization framework for LLM-based workflows and does not by itself guarantee the safety, fairness, privacy, or factual reliability of the underlying models or workflow outputs. If applied to high-stakes domains, such as clinical, legal, or financial decision support, the compiled configurations should be evaluated with domain-specific validation, monitoring, and safeguards. More efficient workflow execution could also lower the cost of deploying harmful or unreliable LLM applications if used irresponsibly. Therefore, FlowCompile should be viewed as an efficiency and configuration-optimization layer rather than a substitute for application-level safety evaluation and responsible deployment practices.

\FloatBarrier


\end{document}